\documentclass{article}

\usepackage[preprint]{nips_2018}

\usepackage{rotating} 

\usepackage[utf8]{inputenc} 
\usepackage[T1]{fontenc}    
\usepackage{hyperref}       
\usepackage{url}            
\usepackage{booktabs}       
\usepackage{amsfonts}       
\usepackage{nicefrac}       
\usepackage{microtype}      
\usepackage{amsmath}
\usepackage{algorithm}
\usepackage{algpseudocode}
\usepackage{float}
\usepackage{graphicx}
\usepackage{subfig}
\usepackage{placeins}

\allowdisplaybreaks[4]

\usepackage{placeins}
\usepackage{algorithm}

\usepackage{etoolbox}
\usepackage{xr}
\makeatletter  
\def\BState{\State\hskip-\ALG@thistlm}
\makeatother

\makeatletter
\AfterEndEnvironment{algorithm}{\let\@algcomment\relax}
\AtEndEnvironment{algorithm}{\kern2pt\hrule\relax\vskip3pt\@algcomment}
\let\@algcomment\relax
\newcommand\algcomment[1]{\def\@algcomment{\footnotesize#1}}

\renewcommand\fs@ruled{\def\@fs@cfont{\bfseries}\let\@fs@capt\floatc@ruled
	\def\@fs@pre{\hrule height.8pt depth0pt \kern2pt}%
	\def\@fs@post{}%
	\def\@fs@mid{\kern2pt\hrule\kern2pt}%
	\let\@fs@iftopcapt\iftrue}
\makeatother

\newtheorem{assumption}{Assumption}

\title{Transfer Learning for Estimating Causal Effects using Neural Networks}

\newcommand*\samethanks[1][\value{footnote}]{\footnotemark[#1]}
\author{
	Sören R. Künzel\thanks{These authors contributed equally to this work.} \\
	UC Berkeley\\
	\texttt{srk@berkeley.edu} \\
	\And
	Bradly C. Stadie\samethanks \\
	UC Berkeley\\
	\texttt{bstadie@berkeley.edu} \\
	\And	
	Nikita Vemuri \\
	UC Berkeley \\
	\AND
	Varsha Ramakrishnan \\
	UC Berkeley \\
	\And
	Jasjeet S. Sekhon \\ 
	UC Berkeley \\
	\texttt{sekhon@berkeley.edu} \\
	\And
    Pieter Abbeel \\
	UC Berkeley \\
	\texttt{pabbeel@cs.berkeley.edu} \\
}

\begin{document}
	\maketitle
	\begin{abstract}
		We develop new algorithms for estimating heterogeneous treatment effects, combining recent developments in transfer learning for neural networks with insights from the causal inference literature. By taking advantage of transfer learning, we are able to efficiently use different data sources that are related to the same underlying causal mechanisms. We compare our algorithms with those in the extant literature using extensive simulation studies based on large-scale voter persuasion experiments and the MNIST database. Our methods can perform an order of magnitude better than existing benchmarks while using a fraction of the data.
	\end{abstract}
	
	\section{Introduction} \label{section:Introduction}

The rise of massive datasets that provide fine-grained information about human beings and their behavior provides unprecedented opportunities for evaluating the effectiveness of treatments. Researchers want to exploit these large and heterogeneous datasets, and they often seek to estimate how well a given treatment works for individuals conditioning on their observed covariates. This problem is important in medicine (where it is sometimes called personalized medicine) \citep{Henderson2016,powers2018some}, digital experiments \citep{taddy2016nonparametric}, economics \citep{Athey2016}, political science \citep{green2012modeling}, statistics
\citep{tian2014simple}, and many other fields. A large number of articles are being written on this topic, but many outstanding questions remain. We present the first paper that applies transfer learning to this problem.


In the simplest case, treatment effects are estimated by splitting a training set into a treatment and a control group. The treatment group receives the treatment, while the control group does not. The outcomes in those groups are then used to construct an estimator for the Conditional Average Treatment Effect (CATE), which is defined as the expected outcome under treatment minus the expected outcome under control given a particular feature vector \citep{athey2015machine}. This is a challenging task because, for every unit, we either observe its outcome under treatment or control, but never both. Assumptions, such as the random assignment of treatment and additional regularity conditions, are needed to make progress. Even with these assumptions, the resulting estimates are often noisy and unstable because the CATE is a vector parameter. Recent research has shown that it is important to use estimators which consider both treatment groups simultaneously 
(\cite{kunzel2017meta, wager2015estimation, nie2017learning, hill2011bayesian}). Unfortunately, these recent advances are often still insufficient to train robust CATE estimators because of the large sample sizes required when the number of covariates is not small.

However, researchers usually fail to use ancillary datasets that are available to them in applications. This is surprising, given the need for additional data to estimate CATE reliably. These ancillary datasets are related to the causal mechanism under investigation, but they are also partially distinct so they cannot be pooled naively, which explains why researches often do not use them. Examples of such ancillary datasets include observations from: experiments in different locations on different populations, different treatment arms, different outcomes, and non-experimental observational studies. The key idea underlying our contributions is that one can substantially improve CATE estimators by transferring information from other data sources.

\textbf{Our contributions are as follows:}

\begin{enumerate}

\item \textbf{We introduce the new problem of transfer learning for estimating heterogeneous treatment effects.}

\item \textbf{We develop the Y-learner for CATE estimation.} We consider the problem of CATE estimation with deep neural networks. We propose the Y-Learner, a CATE estimator designed from the ground up to take advantage of deep neural networks' ability to easily share information across layers. The Y-Learner often achieves state-of-the-art performance on CATE estimation. The Y-learner does not use transfer learning.



\item \textbf{MLRW Transfer for CATE Estimation} adapts the idea of meta-learning regression weights (MLRW) to CATE estimation. Using these learned weights, regression problems can be optimized much more quickly than with random initializations. Though a variety of MLRW algorithms exist, it is not immediately obvious how one should use these methods for CATE estimation. The principle difficulty is that CATE estimation requires the simultaneous estimation of outcomes under both treatment and control, when we only observe one of the outcomes for any individual unit. However, most MLRW transfer methods optimize on a per-task basis to estimate a single quantity. We show that one can overcome this problem with clever use of the Reptile algorithm \citep{reptile}.
While adapting Reptile to work with our problem, we discovered a slight modification to the original algorithm. To distinguish this modification, we refer to it in this paper as \textbf{SF Reptile}, Slow-Fast Reptile. 

\item \textbf{We provide several additional methods for transfer learning for CATE estimation:} warm start, frozen-features, multi-head, and joint training.

\item \textbf{We apply our methods to difficult data problems and show that they perform better than existing benchmarks.} We reanalyze a set of large field experiments that evaluate the effect of a mailer on voter turnout in the 2014 U.S. midterm elections \citep{gerber2017generalizability}. This includes 17 experiments with 1.96 million individuals in total. We also simulate several randomized controlled trials using image data of handwritten digits found in the MNIST database \citep{lecun1998mnist}. We show that our methods, \textbf{MLRW} in particular, obtain better than state-of-the-art performance in estimating CATE, and that they require far fewer observations than extant methods.

\item \textbf{We provide open source code for our algorithms.}\footnote{The software will be released once anonymity is no longer needed. We can also provide an anynomized copy to reviewers upon request.} 
\end{enumerate}

	\newcommand{\Pcal}{\mathcal{P}}
\newcommand{\R}{\mathbb{R}}
\newcommand{\Yobs}{Y^{obs}}
\newcommand{\E}{\mathbb{E}}
\newcommand{\define}{:=}
\renewcommand{\P}{\mathbb{P}}
\newcommand{\emin}{e_{\mbox{\small min}}}
\newcommand{\emax}{e_{\mbox{\small max}}}

\section{CATE ESTIMATION}

We begin by formally introducing the CATE estimation problem. 
Following the potential outcomes framework \citep{rubin1974estimating}, assume there exists a single experiment wherein we observe $N$ i.i.d. distributed units from some super population, $(Y_i(0), Y_i(1), X_i, W_i) \sim \Pcal$. $Y_i(0) \in \R$ denotes the potential outcome of unit $i$ if it is in the control group, $Y_i(1)\in \R$ is the potential outcome of $i$ if it is in the treatment group, $X_i \in \R^d$ is a $d$-dimensional feature vector, and $W_i \in \{0,1\}$ is the treatment assignment.
For each unit in the treatment group ($W_i = 1$), we only observe the outcome under treatment, $Y_i(1)$. For each unit under control ($W_i = 0$), we only observe the outcome under control. Crucially, there cannot exist overlap between the set of units for which $W_i = 1$ and the set for which $W_i = 0$. It is impossible to observe both potential outcomes for any unit. This is commonly referred to as the fundamental problem of causal inference.

However, not all hope is lost. We can still estimate the Conditional Average Treatment Effect (CATE) of the treatment. Let $x$ be an individual feature vector. Then the CATE of $x$, denoted $\tau(x)$, is defined by 

$$\tau(x) = \E[Y(1) - Y(0) | X = x].$$

Estimating $\tau$ is impossible without making further assumptions on the distribution of $(Y_i(0), Y_i(1), X_i, W_i)$. In particular, we need to place two assumptions on our data.


\begin{assumption}[Strong Ignorability, \cite{rosenbaum1983central}] \label{assumption:StrongIgnorability}
$$
    (Y_i(1), Y_i(0)) \perp W |X.     
$$
\end{assumption}
\begin{assumption}[Overlap] \label{assumption:Overlap}
Define the propensity score of $x$ as, 
$$e(x) \define \P(W = 1 | X = x).$$ 
Then there exists constant $0< \emin, ~\emax < 1$ such that for all $x \in \mbox{Support}(X)$, 
$$0 < \emin < e(x) < \emax < 1.$$
In words, $e(x)$ is bounded away from 0 and 1.
\end{assumption}
Assumption \ref{assumption:StrongIgnorability} ensures that there is no unobserved confounder, a random variable which influences both the probability of treatment and the potential outcomes, which would make the CATE unidentifiable. The assumption is particularly strong and difficult to check in applications. Meanwhile, Assumption \ref{assumption:Overlap} rectifies the situation wherein a certain part of the population is always treated or always in the control group. If, for example, all women were in the control group, one cannot identify the treatment effect for women. Though both assumptions are strong, they are nevertheless satisfied by design in randomized controlled trials. While the estimators we discuss would be sensible in observational studies when the assumptions are satisfied, we warn practitioners to be cautious in such studies, especially when the number of covariates is large \citep{overlapAlex}.

Given these two assumptions, there exist many valid CATE estimators. The crux of these methods is to estimate two quantities: 
the control response function, 
$$
\mu_0(x) = \E[Y(0)|X=x],
$$
and the treatment response function,
$$
\mu_1(x) = \E[Y(1)|X=x].
$$
If we denote our learned estimates as $\hat\mu_0(x)$ and $\hat\mu_1(x)$, then we can form the CATE estimate as the difference between the two
$$
\hat \tau(x) = \hat \mu_1(x) - \hat \mu_0(x).
$$
The astute reader may be wondering why we don't simply estimate $\mu_0$ and $\mu_1$ with our favorite function approximation algorithm at this point and then all go home. After all, we have access to the ground truths $\mu_0$ and $\mu_1$ and the corresponding inputs $x$. In fact, it is commonplace to do exactly that. When people directly estimate $\mu_0$ and $\mu_1$ with their favorite model, we call the procedure a T-learner \citep{kunzel2017meta}. Common choices of models include linear models and random forests, though neural networks have recently been considered \citep{nie2017learning}. 

While it may seem like we've triumphed, the T-learner does have some drawbacks \citep{athey2015machine}. It is usually an inefficient estimator. For example, it will often perform poorly when one can borrow information across the treatment conditions. To overcome these deficiencies, a variety of alternative learners have been suggested. Closely related to the T-learner is the idea of estimating the outcome using all of the features and the treatment indicator, without giving the treatment indicator a
special role \citep{hill2011bayesian}. The predicted CATE for an individual unit is then the difference between the predicted values when the treatment assignment indicator is changed from control to treatment, with all other features held fixed. This is called the S-learner, because it uses a single prediction model.

\begin{figure}
	\centering
	\includegraphics[width=0.5\linewidth]{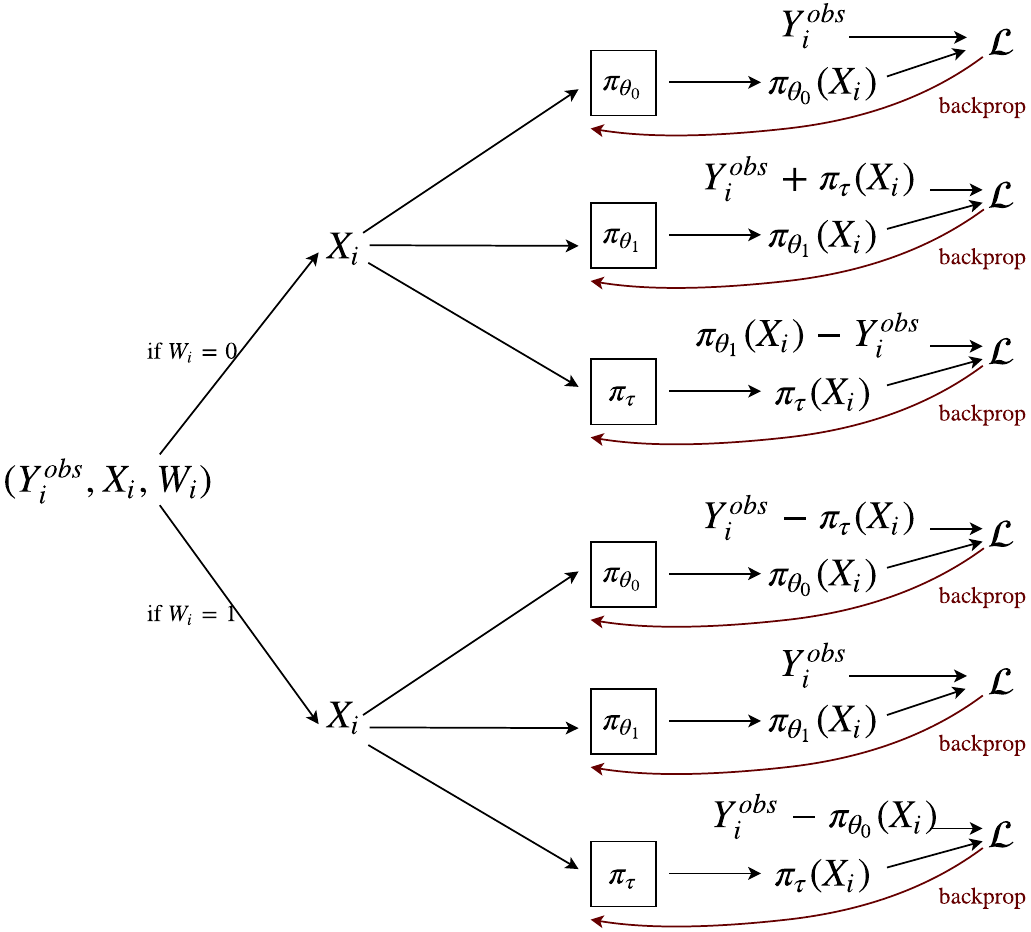}
    \caption{Y-learner with Neural Networks. One of many advanced methods for CATE estimation. See the appendix Section \ref{section:XvsY} and \ref{app:transfer.algos} for a more detailed overview.}
    \label{fig:ylearner}
\end{figure}

In this paper, we suggest another new learner called the $\mathbf{Y}$\textbf{-learner} (See Figure \ref{fig:ylearner}). This learner has been engineered from the ground up to take advantage of some of the unique capabilities of neural networks. See the appendix for a full description of the Y-learner, and additional learners found in the literature. Below, we will use these learners as base algorithms for transfer learning. That is to say, we will use the knowledge gained by training one learner on one experiment to help a new learner with a new underlying experiment train faster with less data. 

\FloatBarrier

\section{Transfer Learning}

\subsection{Background}
The key idea in transfer learning is that new experiments should transfer insights from previous experiments rather than starting learning anew. The most straightforward example of transfer comes from computer vision \citep{finetune0, finetune2, finetune1, donahu}. 
Here, it is standard practice to train a neural network $\pi_\theta$ for one task and then use the trained network weights $\theta$ as initialization for a new task. The hope is that some basic low-level features of a vision system should be quite general and reusable. Starting optimization from networks that have already learned these general features should be faster than starting from scratch. 

Despite its promise, fine-tuning often fails to produce initializations that are uniformly good for solving new tasks \citep{maml}. One potent fix to this problem is a class of algorithms that seek to optimize meta-learning initialization weights \citep{maml, reptile}. In these algorithms, one meta-optimizes over many experiments to obtain neural network weights that can quickly find solutions to new experiments. We will use the Reptile algorithm to learn initialization weights for CATE estimation.

\subsection{Transfer Learning CATE Estimators}

In this section, we consider a scenario wherein one has access to many related causal inference experiments. Across all experiments, the input space $X$ is the same. Let $i$ index an experiment. Each experiment has its own distinct outcome when treatment is received, $\mu^i_1(x)$, and when no treatment is received, $\mu^i_0(x)$. Together, these quantities define the CATE $\tau^i = \mu^i_1 - \mu^i_0$, which we want to estimate. We are usually interested in estimating the CATE by using $X$ to predict $\mu_0^i$ and $\mu_1^i$. However, in transfer learning, the hope is that we can transfer knowledge between experiments such that being able to predict $\mu_0^i, \mu_1^i,$ and $\tau^i$ from experiment $i$ accurately will help us predict $\mu_0^j, \mu_1^j$, and $\tau^j$ from experiment $j$. 

Below, let $\pi_{\theta}$ be a generic expression for a neural network parameterized by $\theta$. Sometimes, parameters will have a subscript indicating if their neural network predicts treatment or control (0 for control and 1 for treatment). Parameters may also have a superscript indicating the experiment number whose outcome is being predicted. For example, $\pi_{\theta_0^2}(x)$ predicts $\mu^2_0(x)$, the outcome under control for Experiment $2$. We will sometimes drop the superscript $i$ when the meaning is clear. All of the transfer algorithms described here are presented in detail in Appendix \ref{app:transfer.algos}.

\textbf{Warm start (also known as fine-tuning):} Experiment $0$ predicts $\pi_{\theta^0_0}(x) = \hat{\mu}^0_0(x)$ and $\pi_{\theta^0_1}(x) = \hat{\mu}^0_1(x)$ to form the CATE estimator $\hat{\tau} = \hat{\mu}^0_1(x) - \hat{\mu}^1_0(x)$. Suppose $\theta^0_0$, $\theta^0_1$ are fully trained and produce a good CATE estimate. For experiment 1, the input space $X$ is identical to the input space for experiment $0$, but the outcomes $\mu^1_0(x)$ and $\mu^1_1(x)$ are different. 
However, we suspect the underlying data representations learned by $\pi_{\theta^0_0}$ and $\pi_{\theta^0_1}$ are still useful. Hence, rather than randomly initialize $\theta_0^1$ and $\theta_1^1$ for experiment 1, we set $\theta_0^1 = \theta_0^0$ and $\theta_1^1 = \theta_1^0$. We then train $\pi_{\theta_0^1}(x) = \hat{\mu}^1_0(x)$ and $\pi_{\theta_1^1}(x) = \hat{\mu}^1_1(x)$. See Figure \ref{fig:warm-start} and Algorithm \ref{alg:warm-t} in the appendix.


\begin{figure}[t]
\begin{center}
\includegraphics[width = 1\textwidth]{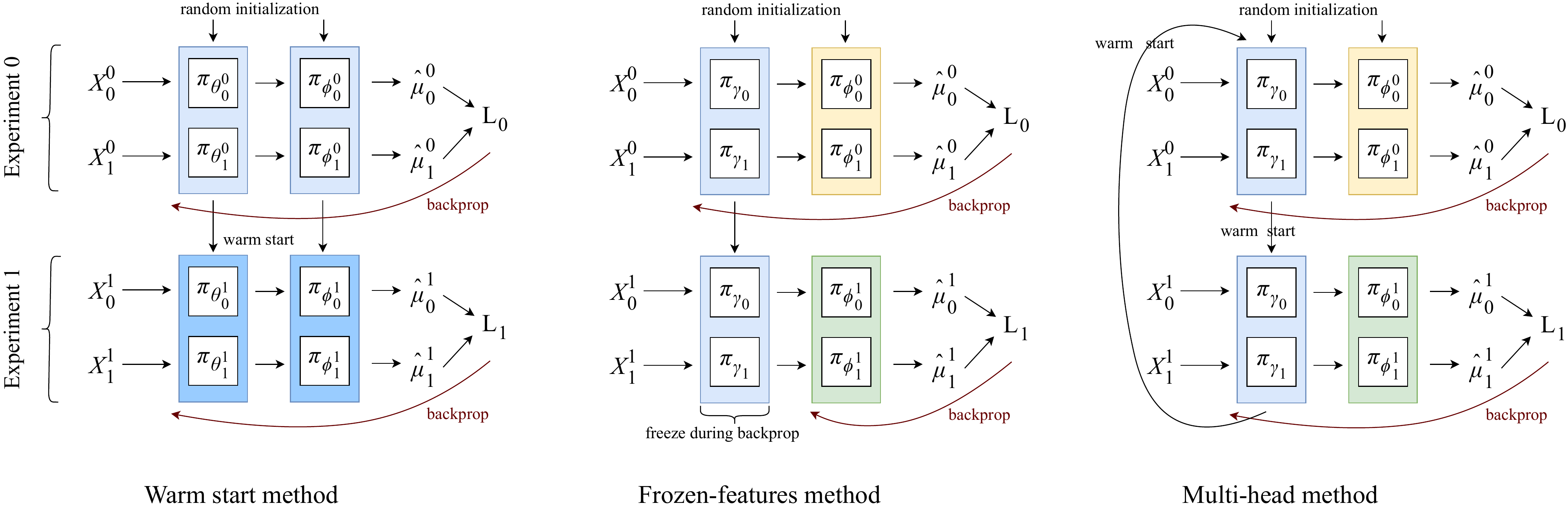}
\end{center}
\caption{Warm start, frozen-features, and multi-head methods for CATE transfer learning. For these figures, we use the T-learner as the base learner for simplicity. All three methods attempt to reuse neural network features from previous experiments. See the appendix for an illustration of joint-training.}
\label{fig:warm-start}
\end{figure} 


\textbf{Frozen-features:} Begin by training $\pi_{\theta^0_0}$ and $\pi_{\theta^0_1}$ to produce good CATE estimates for experiment $0$. Assuming $\theta^0_0$ and $\theta^0_1$ have more than $k$ layers, let $\gamma_0$ be the parameters corresponding to the first $k$ layers of $\theta^0_0$. Define $\gamma_1$ analogously. Since we think the features encoded by $\pi_{\gamma_i} (X)$ would make a more informative input than the raw features $X$, we want to use those features as a transformed input space for $\pi_{\theta_0^1}$ and $\pi_{\theta_1^1}$. To wit, set $z_0 =\pi_{\gamma_0}(x)$ and $z_1 = \pi_{\gamma_1}(x)$. Then form the estimates $\pi_{\theta^1_0} (z_0) = \hat{\mu}^1_0$ and $\pi_{\theta_1^1}(z_1) = \hat{\mu}^1_1$. During training of experiment $1$, we only backpropagate through $\theta_0^1$, $\theta_1^1$ and not through the features we borrowed from $\theta_0^0$ and $\theta_1^0$. See Figure \ref{fig:warm-start} and Algorithm \ref{alg:frozen-t} in the appendix.


\textbf{Multi-head:} In this setup, all experiments share base layers that are followed by experiment-specific layers. The intuition is that the base layers should learn general features, and the experiment-specific layers should transform those features into estimates of $\mu_j^i$. More concretely, let $\gamma_0$ and $\gamma_1$ be shared base layers. Set $z_0 = \pi_{\gamma_0}(x_0)$ and $z_1 = \pi_{\gamma_1}(x_1)$. The base layers are followed by experiment-specific layers $\phi_0^i$ and $\phi_1^i$. Let $\theta_j^i = \left[\gamma_j, \phi_j^i \right]$. Then $\pi_{\theta_j^i}(x) = \pi_{\phi_j^i} \left(\pi_{\gamma_j}(x) \right) = \pi_{\phi_j^i} (z_j) = \hat{\mu}_j^i$. Training alternates between experiments: each $\theta_0^i$ and $\theta_1^i$ is trained for some small number of iterations, and then the experiment and head being trained are switched. Every head is usually trained several times. See Figure \ref{fig:warm-start} Algorithm \ref{alg:multi-t} in the appendix.


\textbf{Joint training:} All predictions share base layers $\theta$. From these base layers, there are two heads per-experiment $i$: one to predict $\mu_0^i$ and one to predict $\mu_1^i$. Every head and the base features are trained simultaneously by optimizing with respect to the loss function $\mathcal{L} = \sum_i  \| \left(\hat{\mu}_0^i - \mu_0^i \right) \| + \| \left(\hat{\mu}_1^i - \mu_1^i \right) \|$ and minimizing over all weights. This will encourage the base layers to learn generally applicable features and the heads to learn features specific to predicting a single $\mu_j^i$. See Figure Algorithm \ref{alg:joint-training}.



\textbf{SF Reptile transfer for CATE estimators:} Similarly to fine-tuning, we no longer provide each experiment with its own weights. Instead, we use data from all experiments to learn weights $\theta_0$ and $\theta_1$, which are good initializers. By good initializers, we mean that starting from $\theta_0$ and $\theta_1$, one can train neural networks $\pi_{\theta_0}$ and $\pi_{\theta_1}$ to estimate $\mu^i_0$ and $\mu^i_1$ for any arbitrary experiment much faster and with less data than starting from random initializations. To learn these good initializations, we use a transfer learning technique called Reptile. The idea is to perform experiment-specific inner updates $U(\theta)$ and then aggregate them into outer updates of the form $\theta_{\text{new}} = \mathbf{\epsilon} \cdot U(\theta) + (1- \mathbf{\epsilon}) \cdot \theta$. In this paper, we consider a slight variation of Reptile. In standard Reptile, $\epsilon$ is either a scalar or correlated to per-parameter weights furnished via SGD. For our problem, we would like to encourage our network layers to learn at different rates. The hope is that the lower layers can learn more general, slowly-changing features like in the frozen features method, and the higher layers can learn comparatively faster features that more quickly adapt to new tasks after ingesting the stable lower-level features. To accomplish this, we take the path of least resistance and make $\epsilon$ a vector which assigns  a different learning rate to each neural network layer. Because our intuition involves slow and fast weights, we will refer to this modification in this paper as SF Reptile: Slow Fast Reptile. Though this change is seemingly small, we found it boosted performance on our problems. See Figure \ref{fig:joint_training} and Algorithm \ref{alg:sf-reptile-t}.

\textbf{MLRW transfer for CATE estimation:} In this method, there exists one single set of weights $\theta$. There are no experiment-specific weights. Furthermore, we do not use separate networks to estimate $\mu_0$ and $\mu_1$. Instead, $\pi_\theta$ is trained to estimate one $\mu_j^i$ at a time. We train $\theta$ with SF Reptile so that in the future $\pi_\theta$ requires minimal samples to fit $\mu_j^i$ from any experiment. To actually form the CATE estimate, we use a small number of training samples to fit $\pi_\theta$ to $\mu_0^i$ and then a small number of training samples to fit $\pi_\theta$ to $\mu_1^i$. We call $\theta$ \textbf{meta-learned regression weights (MLRW)} because they are meta-learned over many experiments to quickly regress onto any $\mu_j^i$. The full MLRW algorithm is presented as Algorithm \ref{alg:sf-reptile-meta-regression}.

\section{Evaluation} \label{section:Simulation}
We evaluate our transfer learning estimators on both real and simulated data.  In our data example, we consider the important problem of voter encouragement. Analyzing a large data set of 1.96 million potential voters, we show how transfer learning across elections and geographic regions can dramatically improve our CATE estimators. This example shows that transfer learning can substantially improve the performance of CATE estimators. \textbf{To the best of our knowledge, this is the first successful demonstration of transfer learning for CATE estimation.} The simulated data has been intentionally chosen to be different in character from our real-world example. In particular, the simulated input space is images and the estimated outcome variable is continuous. 





\subsection{GOTV Experiment} \label{section:GOTV}



To evaluate transfer learning for CATE estimation on real data, we reanalyze a set of large field experiments with more than 1.96 million potential voters \citep{gerber2017generalizability}. The authors conducted 17 experiments to evaluate the effect of a mailer on voter turnout in the 2014 U.S.\ Midterm Elections. 
The mailer informs the targeted individual whether or not they voted in the past four major elections (2006, 2008, 2010, and 2012), and it compares their voting behavior with that of the people in the same state. The mailer finishes with a reminder that their voting behavior will be monitored. The idea is that social pressure---i.e., the social norm of voting---will encourage people to vote. 
The likelihood of voting increases by about 2.2\% (s.e.=0.001) when given the mailer.

Each of the experiments target a different state. This results in different populations, different ballots, and different electoral environments. In addition to this, the treatment is slightly different in each experiment, as the median voting behavior in each state is different.
However, there are still many similarities across the experiments, so there should be gains from transferring information. 

In this example, the input $X$ is a voter's demographic data including age, past voting turnout in 2006, 2008, 2009, 2010, 2011, 2012, and 2013, marital status, race, and gender. The treatment response function $\hat{\mu}_1(x)$ estimates the voting propensity for a potential voter who receives a mailer encouraging them to vote. The control response function $\hat{\mu}_0$ estimates the voting propensity if that voter did not receive a mailer. 
The CATE $\tau$ is thus the change in the probability of voting when a unit receives a mailer. The complete dataset has this data over 17 different states. Treating each state as a separate experiment, we can perform transfer learning across them.  

\begin{center}
	\begin{tabular}{| p{2.3cm} | p{2.3cm} | p{2.3cm} |p{2.3cm} |p{2.3cm} |}
		\hline
		$x$ & outcome & $\mu_0$ & $\mu_1$ & $\tau$ \\ [0.5ex] 
		\hline \hline
		a voter profile & \parbox[t]{2cm}{The voter's \\ propensity to \\ vote} & \parbox[t]{2cm}{The voter's \\ propensity to \\ vote when they \\ do not receive \\ a mailer} & \parbox[t]{2cm}{The voter's \\ propensity to \\ vote when they \\ do receive a \\ mailer} & \parbox[t]{2cm}{Change in the \\ voter's \\ propensity to \\ vote after \\ receiving a \\ mailer \\ } \\  \hline
		\hline
	\end{tabular}
\end{center}

Being able to estimate the treatment effect of sending a mailer is an important problem in elections. We may wish to only treat people whose likelihood of voting would significantly increase when receiving the mailer, to justify the cost for these mailers. Furthermore, we wish to avoid sending mailers to voters who will respond negatively to them. 
This negative response has been previously observed and is therefore feasible and a relevant problem---e.g., some recipients call their Secretary of State's office or local election registrar to complain \citep{mann2010there,michelson2016risk}.  

\subsubsection*{Evaluating CATE estimators on real data}
Evaluating a CATE estimator on real data is difficult since one does not observe the true CATE or the individual treatment effect, $Y_i(1) - Y_i(0)$, for any unit because by definition only one of the two outcomes is observed for any unit. One could use the original features and simulate the outcome features, but this would require us to create a response model. Instead, we estimate the "truth" on the real data using linear models (version 1) or random forests (version 2), and we then draw the data based on these estimates. For a detailed description, we refer to Appendix \ref{section:Appendix-GOTV}. We then ask the question: how do the various methods perform when they have less data than the entire sample?

\begin{figure}
	\centering
	\includegraphics[width=.8\linewidth]{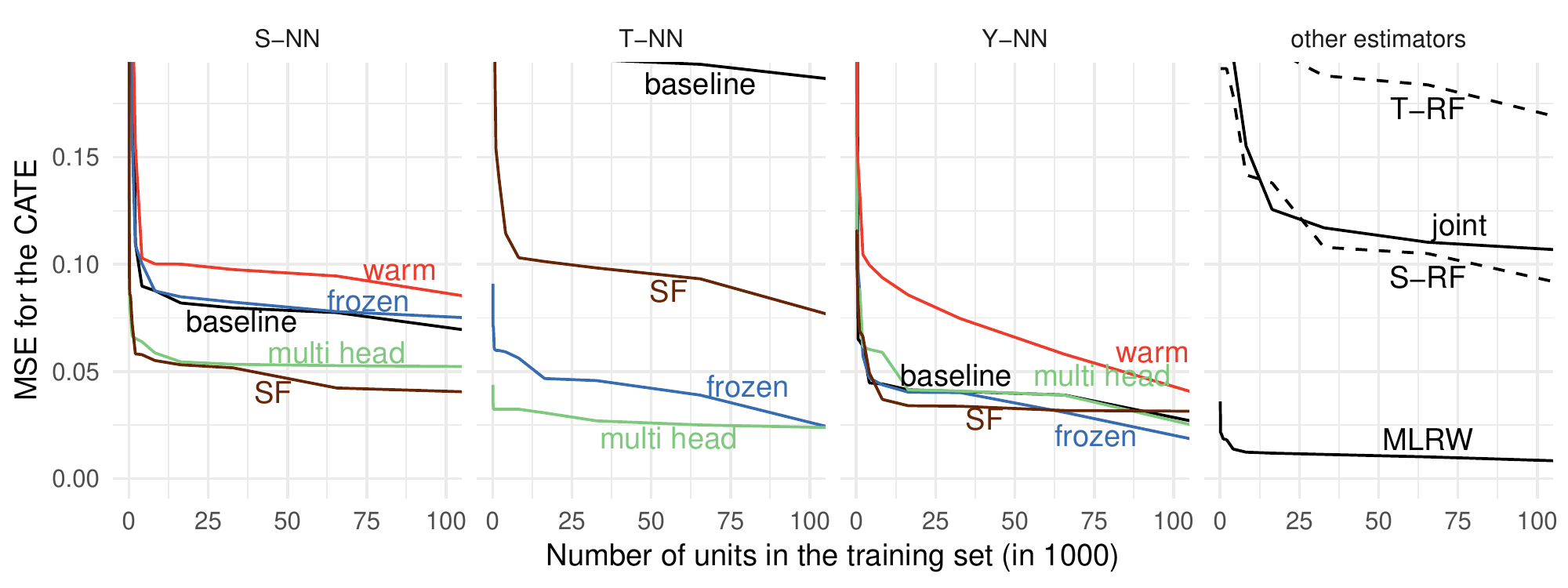}
    \caption{Social Pressure and Voter Turnout, Version 1. Our results far exceed the previous state of the art, which are represented here as S-RF, T-RF, and the baseline method for S-NN and T-NN. Our new methods are Y-NN and the transfer learning methods: warm, frozen, multi-head, joint, SF Reptile, and MLRW.}
    	\label{fig:mainpapergotv21lm1}
\end{figure}

\begin{figure}
	\centering
	\includegraphics[width=.8\linewidth]{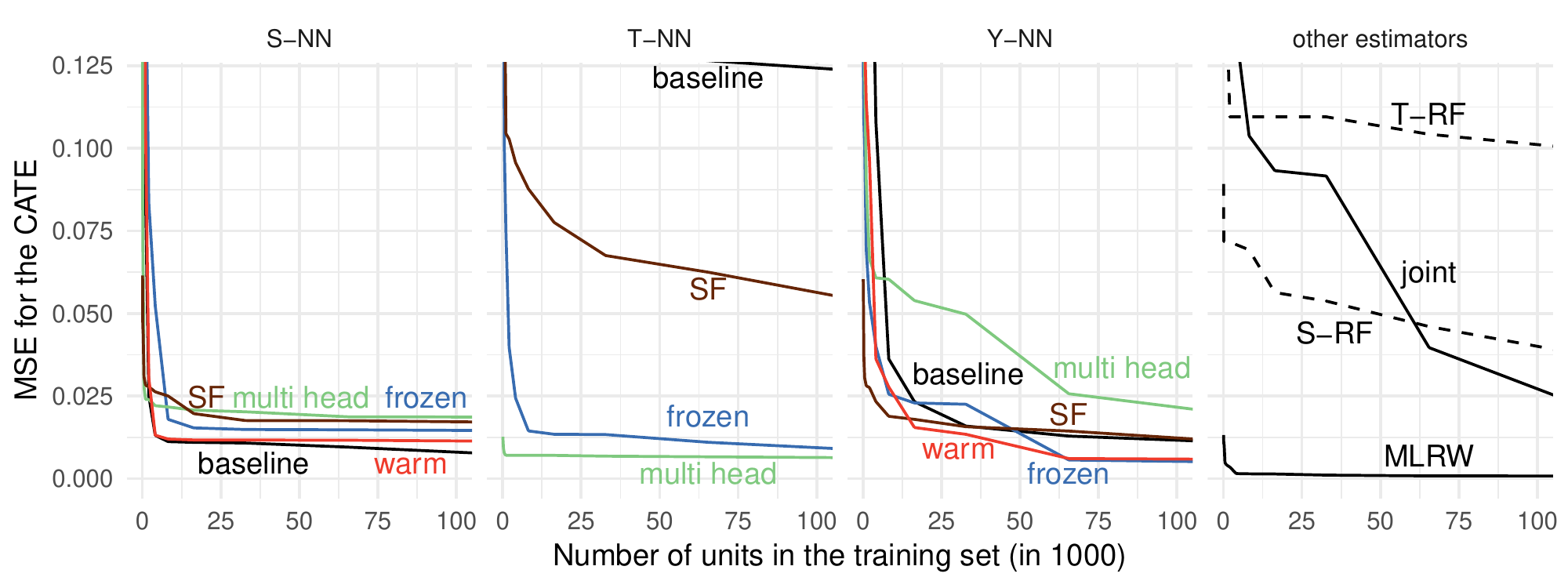}
       \caption{Social Pressure and Voter Turnout, Version 2. Our results exceed the previous state of the art results, which are represented here as S-RF, T-RF, and the baseline method for S-NN and T-NN. Our new methods are Y-NN and the transfer learning methods: warm, frozen, multi-head, joint, SF Reptile, and MLRW.}
    	\label{fig:mainpaperGOTV22rf1}
\end{figure}

We evaluate S-NN, T-NN, and Y-NN using our transfer learning methods. We also added a baseline benchmark which does not use any transfer learning for each of the CATE estimators. 
In addition to this, we added the S-RF and T-RF as random forest baselines, as well as the Joint estimator and the MLRW estimator, both of which use transfer learning.
Figure \ref{fig:mainpapergotv21lm1} shows the performance of these estimators when the regression functions were created using a linear model, and Figure \ref{fig:mainpaperGOTV22rf1} shows the same, but the response functions are created using a random forest fitted on the real data. 

In previous work, the non-transfer tree-based estimators such as T-RF and S-RF have achieved state of the art results on this problem \citep{kunzel2017meta}. For CATE estimation, these methods are very competitive baselines \citep{green2012modeling}. Happily for us, even non-transfer neural-network-based learners vastly outperform the prior art. In both examples, non-transfer S-NN, T-NN, and Y-NN learners are better or not much worse than T-RF and S-RF. S-NN and Y-NN perform extremely well in this example. Better still, our transfer learning approaches consistently outperform all classical baselines and non-transfer neural network learners on this benchmark. Positive transfer between experiments is readily apparent. 

We find that multi-head, frozen features, and SF are usually the best methods to improve an existing neural network-based CATE estimator.  \textbf{The best estimator is MLRW.} This algorithm consistently converges to a very good solution with very few observations.




\subsection{MNIST Example} \label{section:MNIST}


In the previous experiment, we observed that the MLRW estimator performed most favorably and transfer learning significantly improved upon the baseline. To confirm that this conclusion is not specific to voter persuasion studies, we consider in this section intentionally a very different type of data.
Recently, \cite{nie2017learning} introduced a simulation study wherein MNIST digits are rotated by some number of degrees $\alpha$; with $\alpha$ furnished via a single data generating process that depends on the value of the depicted digit. They then attempt to do CATE estimation to measure the heterogeneous treatment effect of a digit's label. 

Motivated by this example, we develop a data generating process using MNIST digits wherein transfer learning for CATE estimation is applicable. In our example, the input $X$ is an MNIST image. We have $k$ data generating processes which return different outcomes for each input when given either treatment or control. Thus, under some fixed data generating process, $\mu_0$ represents the outcome when the input image $X$ is given the control, $\mu_1$ represents the outcome when $X$ is given the treatment, and $\tau$ is the difference in outcomes given the placement of $X$ in the treatment or control group. Each data generating process has different response functions ($\mu_0$ and $\mu_1$) and thus different CATEs ($\tau$), but each of these functions only depend on the image label presented in the image $X$. We thus hope that transfer learning could expedite the process of learning features which are indicative of the label. See Appendix \ref{app:simulation} for full details of the data generation process. 
In Figure \ref{fig:MNIST} of Appendix \ref{app:simulation}, we confirm that a transfer learning strategy outperforms its non-transfer learning counterpart, even on image data, and also that MLRW performs well.

	\section{Discussion and Conclusion}



In this paper, we proposed the problem of transfer learning for CATE estimation. One immediate question the reader may be left with is why we chose the transfer learning techniques we did. We only considered two common types of transfer: 
(1) Basic fine tuning and weights sharing techniques common in the computer vision literature \citep{finetune0,finetune2,finetune1, donahu, koch},
(2) Techniques for learning an initialization that can be quickly optimized \citep{maml,hugo,reptile}. However, many further techniques exist. Yet, transfer learning is an extensively studied and perennial problem \citep{schmid,bengio,thurn2,thrun,taylor,silver}. In \cite{oriol}, the authors attempt to combine feature embeddings that can be utilized with non-parametric methods for transfer. \cite{proto} is an extension of this work that modifies the procedure for sampling examples from the support set during training. \cite{marcin} and related techniques try to meta-learn an optimizer that can more quickly solve new tasks. \cite{progressive} attempts to overcome forgetting during transfer by systematically introducing new network layers with lateral connections to old frozen layers. \cite{yu} uses networks with memory to adapt to new tasks. We invite the reader to review \cite{maml} for an excellent overview of the current transfer learning landscape. Though the majority of the discussed techniques could be extended to CATE estimation, our implementations of \cite{progressive, marcin} proved difficult to tune and consequently learned very little. Furthermore, we were not able to successfully adapt \cite{proto} to the problem of regression. We decided to instead focus our attention on algorithms for obtaining good initializations, which were easy to adapt to our problem and quickly delivered good results without extensive tuning. 




	\clearpage
	
	\bibliographystyle{apalike}
	\bibliography{main.bbl}

\begin{thebibliography}{}

\bibitem[Andrychowicz et~al., 2016]{marcin}
Andrychowicz, M., Denil, M., Gomez, S., Hoffman, M.~W., Pfau, D., Schaul, T.,
  and de~Freitas, N. (2016).
\newblock Learning to learn by gradient descent by gradient descent.
\newblock {\em Neural Information Processing Systems (NIPS)}.

\bibitem[Athey and Imbens, 2015]{athey2015machine}
Athey, S. and Imbens, G.~W. (2015).
\newblock Machine learning methods for estimating heterogeneous causal effects.
\newblock {\em stat}, 1050(5).

\bibitem[Athey and Imbens, 2016]{Athey2016}
Athey, S. and Imbens, G.~W. (2016).
\newblock {Recursive partitioning for heterogeneous causal effects}.
\newblock {\em Proceedings of the National Academy of Sciences of the United
  States of America}, 113(27):7353--60.

\bibitem[Bengio et~al., 1992]{bengio}
Bengio, S., Bengio, Y., Cloutier, J., and Gecsei, J. (1992).
\newblock On the optimization of a synaptic learning rule.
\newblock {\em Biological Neural Networks}.

\bibitem[Bourdev et~al., 2011]{finetune1}
Bourdev, L., Maji, S., and Malik, J. (2011).
\newblock Describing people: A poselet-based approach to attribute
  classification.
\newblock {\em ICCV}.

\bibitem[D'Amour et~al., 2017]{overlapAlex}
D'Amour, A., Ding, P., Feller, A., Lei, L., and Sekhon, J. (2017).
\newblock Overlap in observational studies with high-dimensional covariates.
\newblock {\em arXiv preprint arXiv:1711.02582}.

\bibitem[Donahue et~al., 2014]{donahu}
Donahue, J., Jia, Y., Vinyals, O., Hoffman, J., Zhang, N., Tzeng, E., and
  Darrell, T. (2014).
\newblock Decaf: A deep convolutional activation feature for generic visual
  recognition.
\newblock {\em International Conference on Machine Learning (ICML)}.

\bibitem[Finn et~al., 2017]{maml}
Finn, C., Abbeel, P., and Levine, S. (2017).
\newblock Model-agnostic metalearning for fast adaptation of deep networks.
\newblock {\em ICML}.

\bibitem[Gerber et~al., 2017]{gerber2017generalizability}
Gerber, A.~S., Huber, G.~A., Fang, A.~H., and Gooch, A. (2017).
\newblock The generalizability of social pressure effects on turnout across
  high-salience electoral contexts: Field experimental evidence from 1.96
  million citizens in 17 states.
\newblock {\em American Politics Research}, 45(4):533--559.

\bibitem[Green and Kern, 2012]{green2012modeling}
Green, D.~P. and Kern, H.~L. (2012).
\newblock Modeling heterogeneous treatment effects in survey experiments with
  bayesian additive regression trees.
\newblock {\em Public opinion quarterly}, 76(3):491--511.

\bibitem[Henderson et~al., 2016]{Henderson2016}
Henderson, N.~C., Louis, T.~A., Wang, C., and Varadhan, R. (2016).
\newblock Bayesian analysis of heterogeneous treatment effects for
  patient-centered outcomes research.
\newblock {\em Health Services and Outcomes Research Methodology},
  16(4):213--233.

\bibitem[Hill, 2011]{hill2011bayesian}
Hill, J.~L. (2011).
\newblock Bayesian nonparametric modeling for causal inference.
\newblock {\em Journal of Computational and Graphical Statistics},
  20(1):217--240.

\bibitem[Koch, 2015]{koch}
Koch, G. (2015).
\newblock Siamese neural networks for one-shot image recognition.
\newblock {\em ICML Deep Learning Workshop}.

\bibitem[K{\"u}nzel et~al., 2017]{kunzel2017meta}
K{\"u}nzel, S., Sekhon, J., Bickel, P., and Yu, B. (2017).
\newblock Meta-learners for estimating heterogeneous treatment effects using
  machine learning.
\newblock {\em arXiv preprint arXiv:1706.03461}.

\bibitem[LeCun, 1998]{lecun1998mnist}
LeCun, Y. (1998).
\newblock The mnist database of handwritten digits.
\newblock {\em http://yann. lecun. com/exdb/mnist/}.

\bibitem[Mann, 2010]{mann2010there}
Mann, C.~B. (2010).
\newblock Is there backlash to social pressure? a large-scale field experiment
  on voter mobilization.
\newblock {\em Political Behavior}, 32(3):387--407.

\bibitem[Michelson, 2016]{michelson2016risk}
Michelson, M.~R. (2016).
\newblock The risk of over-reliance on the institutional review board: An
  approved project is not always an ethical project.
\newblock {\em PS: Political Science \& Politics}, 49(02):299--303.

\bibitem[Munkhdalai and Yu, 2017]{yu}
Munkhdalai, T. and Yu, H. (2017).
\newblock Meta networks.
\newblock {\em ICML}.

\bibitem[Nichol et~al., 2018]{reptile}
Nichol, A., Achiam, J., and Schulman, J. (2018).
\newblock On first-order meta-learning algorithms.
\newblock {\em CoRR}, abs/1803.02999.

\bibitem[Nie and Wager, 2017]{nie2017learning}
Nie, X. and Wager, S. (2017).
\newblock Learning objectives for treatment effect estimation.
\newblock {\em arXiv preprint arXiv:1712.04912}.

\bibitem[Powers et~al., 2018]{powers2018some}
Powers, S., Qian, J., Jung, K., Schuler, A., Shah, N.~H., Hastie, T., and
  Tibshirani, R. (2018).
\newblock Some methods for heterogeneous treatment effect estimation in high
  dimensions.
\newblock {\em Statistics in medicine}.

\bibitem[Ravi and Larochelle, 2017]{hugo}
Ravi, S. and Larochelle, H. (2017).
\newblock Optimization as a model for few-shot learning.
\newblock {\em International Conference on Learning Representations (ICLR)}.

\bibitem[Rosenbaum and Rubin, 1983]{rosenbaum1983central}
Rosenbaum, P.~R. and Rubin, D.~B. (1983).
\newblock The central role of the propensity score in observational studies for
  causal effects.
\newblock {\em Biometrika}, 70(1):41--55.

\bibitem[Rubin, 1974]{rubin1974estimating}
Rubin, D.~B. (1974).
\newblock Estimating causal effects of treatments in randomized and
  nonrandomized studies.
\newblock {\em Journal of educational Psychology}, 66(5):688.

\bibitem[Rusu et~al., 2016]{progressive}
Rusu, A.~A., Rabinowitz, N.~C., Desjardins, G., Soyer, H., Kirkpatrick, J.,
  Kavukcuoglu, K., Pascanu, R., and Hadsell, R. (2016).
\newblock Progressive neural networks.
\newblock {\em CoRR, vol. abs/1606.04671}.

\bibitem[Saenko and Darrell, 2010]{finetune2}
Saenko, K., K. B. F.~M. and Darrell, T. (2010).
\newblock Adapting visual category models to new domains.
\newblock {\em ECCV}.

\bibitem[Schmidhuber, 1992]{schmid}
Schmidhuber, J. (1992).
\newblock Learning to control fast-weight memories: An alternative to dynamic
  recurrent networks.
\newblock {\em Neural Computation}.

\bibitem[Silver et~al., 2013]{silver}
Silver, Yand, and Li (2013).
\newblock Lifelong machine learning systems: Beyond learning algorithms.
\newblock {\em DAAAI Spring Symposium-Technical Report, 2013.}

\bibitem[Snell et~al., 2017]{proto}
Snell, J., ~, Swersky, K., and Zemel, R. (2017).
\newblock Prototypical networks for few-shot learning.
\newblock {\em arXiv preprint arXiv:1703.05175}.

\bibitem[Taddy et~al., 2016]{taddy2016nonparametric}
Taddy, M., Gardner, M., Chen, L., and Draper, D. (2016).
\newblock A nonparametric bayesian analysis of heterogenous treatment effects
  in digital experimentation.
\newblock {\em Journal of Business \& Economic Statistics}, 34(4):661--672.

\bibitem[Taylor and Stone, 2009]{taylor}
Taylor and Stone (2009).
\newblock Transfer learning for reinforcement learning domains: A survey.
\newblock {\em DAAAI Spring Symposium-Technical Report, 2013.}

\bibitem[Thrun, 1996]{thurn2}
Thrun (1996).
\newblock Is learning the n-th thing any easier than learning the first?
\newblock {\em NIPS}.

\bibitem[Thrun and Pratt, 1998]{thrun}
Thrun, S. and Pratt, L. (1998).
\newblock Learning to learn.
\newblock {\em Springer Science and Business Media}.

\bibitem[Tian et~al., 2014]{tian2014simple}
Tian, L., Alizadeh, A.~A., Gentles, A.~J., and Tibshirani, R. (2014).
\newblock A simple method for estimating interactions between a treatment and a
  large number of covariates.
\newblock {\em Journal of the American Statistical Association},
  109(508):1517--1532.

\bibitem[Vinyals et~al., 2016]{oriol}
Vinyals, O., Blundell, C., Lillicrap, T., Wierstra, D., and et~al (2016).
\newblock Matching networks for one shot learning.
\newblock {\em Neural Information Processing Systems (NIPS)}.

\bibitem[Wager and Athey, 2017]{wager2015estimation}
Wager, S. and Athey, S. (2017).
\newblock Estimation and inference of heterogeneous treatment effects using
  random forests.
\newblock {\em Journal of the American Statistical Association}.

\bibitem[Welinder et~al., 2010]{finetune0}
Welinder, P., Branson, S., Mita, T., Wah, C., Schroff, F., Belongie, S., and
  Perona, P. (2010).
\newblock Caltech-ucsd birds 200. technical report cns-tr-2010-001.
\newblock {\em California Institute of Technology}.

\end{thebibliography}
	
    \clearpage

    \appendix
    \clearpage
    \newcommand{\N}{\mathcal{N}}

\section{Appendix: Simulation Studies and Application}
\FloatBarrier
\begin{figure}[H]
	\centering
	\includegraphics[width=.5\linewidth]{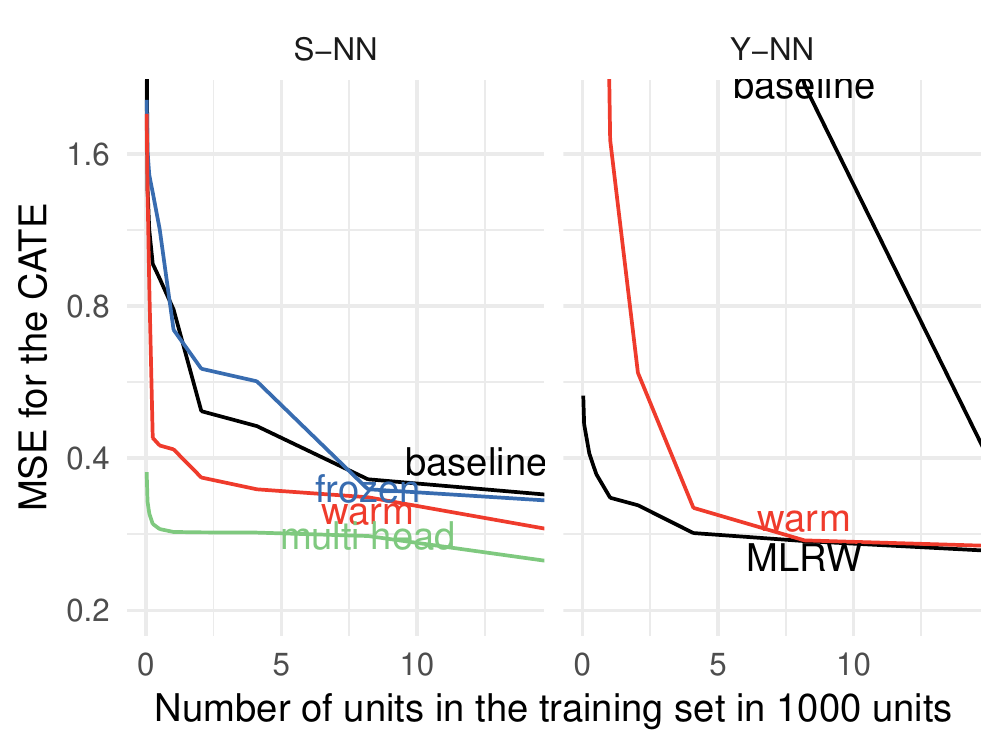}
    \caption{MNIST task}
    	\label{fig:MNIST}
\end{figure}

\label{app:simulation}
\subsection{MNIST Simulation}
For our MNIST simulation study (Section \ref{section:MNIST}), we used the MNIST database \citep{lecun1998mnist} which contains labeled handwritten images. We follow here the notation of \cite{nie2017learning}, who introduce a very similar simulation study which is not trying to evaluate transfer learning for CATE estimation, but instead emulates a RCT with the goal to evaluate different CATE estimators. 

The MNIST data set contains labeled image data $(X_i, C_i)$, where $X_i$ denotes the raw image of $i$ and $C_i \in \{0, \ldots, 9\}$ denotes its label. 
We create $k$ Data Generating Processes (DGPs), $D_1,~\ldots, ~ D_k$, each of which specifies a distribution of $(Y_i(0),Y_i(1), W_i, X_i)$ and represents different CATE estimation problems. 

In this simulation, we let $W_i = 0$ if the image $X_i$ is placed in the control, and $W_i = 1$ if the image $X_i$ is placed in the treatment. $Y_i(W_i)$ quantifies the the outcome of $X_i$ under $W_i$.

To generate a DGP $D_j$ , we first sample weights in the following way, 
\begin{align*}
m^j(0), ~m^j(1), \ldots, ~m^j(9) &\overset{iid}{\sim} \mbox{Unif}(-3,~ 3),\\ 
t^j(0), ~t^j(  1), \ldots, ~t^j(9) &\overset{iid}{\sim} \mbox{Unif}(-1,~ 1),\\ 
p^j(0), ~p^j(1), \ldots, ~p^j(9) &\overset{iid}{\sim} \mbox{Unif}(0.3,~ 0.7),
\end{align*}
and we define the response functions and the propensity score as
\begin{align*}
\mu^j_0(C_i) & = m^j(C_i) + 3 C_i,\\ 
\mu^j_1(C_i) & = \mu^j_0(C_i) + t^j(C_i),\\
e^j(C_i) & = p^j(C_i).
\end{align*}
To generate $(Y_i(0),Y_i(1), W_i, X_i)$ from $D_j$, we fist sample a $(X_i, C_i)$ from the MNIST data set, and we then generate $Y_i(0),Y_i(1)$, and $W_i$ in the following way:
\begin{align*}
\varepsilon_i &\overset{iid}{\sim} \N(0,1)\\
Y_i(0) &= \mu_0(C_i) + \varepsilon_i\\
Y_i(1) &=  \mu_1(C_i) + \varepsilon_i\\
W_i &\sim  \mbox{Bern}(e(C_i)).
\end{align*}

During training, $X_i, W_i$, and $Y_i$ are made available to the convolutional neural network, which then predicts $\hat{\tau}$ given a test image $X_i$ and a treatment $W_i$. $\tau$ is the difference in the outcome given the difference in treatment and control.

Having access to multiple DGPs can be interpreted as having access to prior experiments done on a similar population of images, allowing us to explore the effects of different transfer learning methods when predicting the effect of a treatment in a new image.

    \subsection{GOTV Data Example and Simulation} \label{section:Appendix-GOTV} \label{sec:simulatedGOTV}
In this section, we describe how the simulations for the GOTV example in the main paper were done and we discuss the results of a much bigger simulation study with 51 experiments which is summarized in Tables \ref{table:LM}, \ref{table:RF}, and \ref{table:RF1}.

\subsubsection{Data Generating Processes for Our Real World Example}
For our data example, we took one of the experiments conducted by \cite{gerber2017generalizability}. The study took place in 2014 in Alaska and 252,576 potential voters were randomly assigned in a control and a treatment group. Subjects in the treatment group were sent a mailer as described in the main text and their voting turnout was recorded. 

To evaluate the performance of different CATE estimators we need to know the true CATEs, which are unknown due to the fundamental problem of causal inference.
To still be able to evaluate CATE estimators researchers usually estimate the potential outcomes using some machine learning method and then generate the data from this estimate. This is to some extend also a simulation, but unlike classical simulation studies it is not up to the researcher to determine the data generating distribution.
The only choice of the researcher lies in the type of estimator she uses to estimate the response functions. To avoid being mislead by artifacts created by a particular method, we used a linear model in \textbf{Real World Data Set 1} and random forests estimator in \textbf{Real World Data Set 2}.

Specifically, we generate for each experiment a true CATE and we simulate new observed outcomes based on the real data in four steps.
\begin{enumerate}
\item We first use the estimator of choice (e.g., a random forests estimator) and train it on the treated units and on the control units separately to get estimates for the response functions, $\mu_0$ and $\mu_1$.
\item Next, we sample $N$ units from the underlying experiment to get the features and the treatment assignment of our samples $(X_i, W_i)_{i=1}^N$.
\item We then generate the true underlying CATE for each unit using $\tau_i = \tau(X_i) = \mu_1(X_i) - \mu_0(X_i)$.
\item Finally we generate the observed outcome by sampling a Bernoulli distributed variable around mean $\mu_i$. 
  \begin{align*}
  	&\Yobs_i \sim \mbox{Bern}(\mu_i),&
    &\mu_i = 
    \begin{cases}
    \mu_0(X_i) ~ ~ \mbox{if} ~ ~ W = 0,\\
    \mu_1(X_i) ~ ~ \mbox{if} ~ ~ W = 1.
    \end{cases}&
  \end{align*}
\end{enumerate}

After this procedure, we have 17 data sets corresponding to the 17 experiments for which we know the true CATE function, which we can now use to evaluate CATE estimators and CATE transfer learners.

\subsubsection{Data Generating Processes for Our Simulation Study}
Simulations motivated by real-world experiments are important to assess whether our methods work well for voter persuasion data sets, but it is important to also consider other settings to evaluate the generalizability of our conclusions. 

To do this, we first specify the control response function, $\mu_0(x) = \E[Y(0)|X = x] \in [0,1]$, 
and the treatment response function, $\mu_1(x) = \E[Y(1)|X = x] \in [0,1]$.

We then use each of the 17 experiments to generate a simulated experiment in the following way: 
\begin{enumerate}
\item We sample $N$ units from the underlying experiment to get the features and the treatment assignment of our samples $(X_i, W_i)_{i=1}^N$.
\item We then generate the true underlying CATE for each unit using $\tau_i = \tau(X_i) = \mu_1(X_i) - \mu_0(X_i)$.
\item Finally we generate the observed outcome by sampling a Bernoulli distributed variable around mean $\mu_i$. 
  \begin{align*}
  	&\Yobs_i \sim \mbox{Bern}(\mu_i),&
    &\mu_i = 
    \begin{cases}
    \mu_0(X_i) ~ ~ \mbox{if} ~ ~ W = 0,\\
    \mu_1(X_i) ~ ~ \mbox{if} ~ ~ W = 1.
    \end{cases}&
  \end{align*}
\end{enumerate}

The experiments range in size from 5,000 units to 400,000 units per experiment and the covariate vector is 11 dimensional and the same as in the main part of the paper.
We will present here three different setup. 

\textbf{Simulation LM} (Table \ref{table:LM}):
We choose here $N$ to be all units in the corresponding experiment.
Sample $\beta^0 = (\beta^0_1, \ldots, \beta^0_d) \overset{iid}{\sim} \N(0,1)$ and
$\beta^1 = (\beta^1_1, \ldots, \beta^1_d) \overset{iid}{\sim} \N(0,1)$ and define,
\begin{align*}
\mu_0(x) &= \mbox{logistic}\left(x \beta^0\right), \\
\mu_1(x) &= \mbox{logistic}\left(x \beta^1\right).
\end{align*}
\\ \textbf{Simulation RF} (Table \ref{table:RF}):
We choose here $N$ to be all units in the corresponding experiment.
\begin{enumerate}
	\item Train a random forests estimator on the real data set and define $\mu_0$ to be the resulting estimator,
	\item Sample a covariate $f$ (e.g., age),
    \item ample a random value in the support of $f$ (e.g., 38),
    \item Sample a shift $s \sim \N(0, 4).$
\end{enumerate}
Now define the potential outcomes as follows:
\begin{align*} 
\mu_0(x) &= \mbox{trained Random Forests algorithm}\\
\mu_1(x) &= \mbox{logistic}\left(\mbox{logit}\left(\mu_0(x) + s * 1_{f\ge v} \right)\right)
\end{align*}
\\ \textbf{Simulation RFt} (Table \ref{table:RF1}):
This experiment is the same as Simulation RF, but use only one percent of the data, $N = \frac{\# units}{100}$.

\subsubsection*{Results of 42 Simulations}
Even though we combine each Simulation setup with 17 experiments, we only report the first 14, because the last three don't add any new insight, but they don't fit well on the page. 
Looking at Tables \ref{table:LM}, \ref{table:RF}, and \ref{table:RF1}, we observe that MLRW is the best performing transfer learner. In fact, for Simulation LM it is the best in 8 out of 17 experiments, in Simulation RF it is the best in 11 out of 17 experiments, and in Simulation RFt it is best in 10 out of 17 experiments. We also notice that in cases, where it is not the best performing estimator, it is usually very close to the best and it does not fail terribly anywhere. 
For the other transfer method, we note that frozen features, multi-head, and SF works very well and consistently improves upon the baseline learners which are not using outside information. Warm Start, however, does not work well and often even leads to worse results than the baseline estimators.

    \section{Y-learner} \label{section:XvsY}
\begin{figure}
	\centering
	\includegraphics[width=1\linewidth]{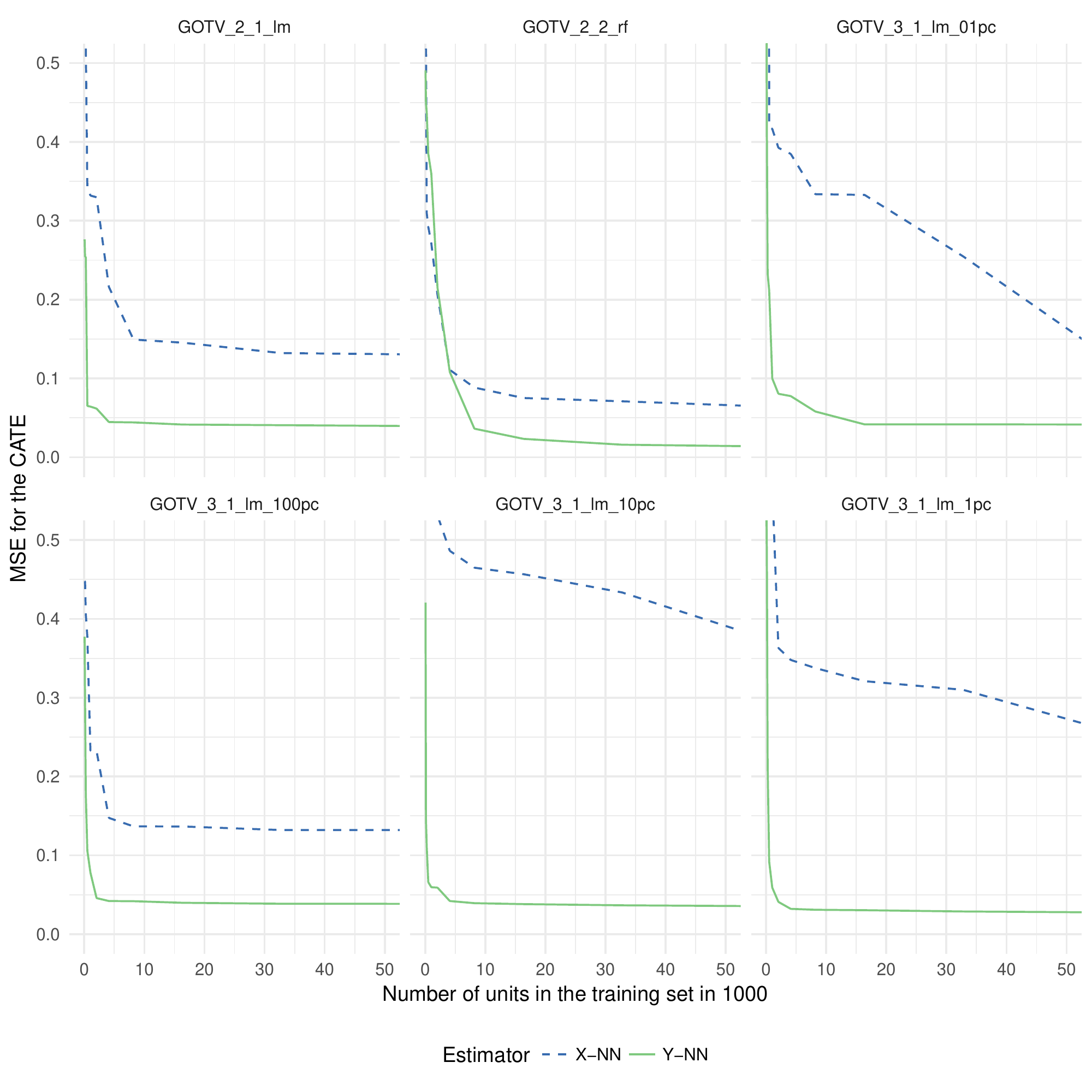}
    \caption{In this figure, we compare the Y and the X learner on six simulated data sets. A precise description on how the data was created can be found in Section \ref{sec:simulatedGOTV}}
    	\label{fig:compareXY}
\end{figure}

In this section, we show the favorable behavior of the Y-learner over the X-learner. In order to show this, we implemented the X-learner exactly as it is described in \citep{kunzel2017meta} and the Y-learner as it is described in Algorithm \ref{algo:Ylearner}. 
Figure \ref{fig:compareXY} shows the MSE in proportion to its sample size. We can see that the X--learner is consistently outperformed on all these data sets by the Y-learner. 
We note that all these data sets were intentionally crated to be very similar to the GOTV data set we are interested in studying. Therefore these data sets are not extremely different from each other, and it is possible that the X-NN performs much better on different data sets. 
\FloatBarrier

\subsubsection*{The Y-Learner}
Another important advantage of neural networks is that they can be trained jointly. This enables us to adapt well-performing meta-learners to perform even better. 
Specifically, we used the idea of X-NN to propose 
a new CATE estimator, which we call Y-NN.\footnote{Y is chosen as it is the next letter in the alphabet after X. However, this is not a meta-learner because there is no obvious way to extend it to arbitrary base learners, such as RF or BART.}
The X-learner is essentially a two step procedure.
In the first stage, the outcome functions, $\hat \mu_0$ and $\hat \mu_1$, are estimated and the individual treatment effects are imputed: 
\begin{align*}
&D_i^1 \define Y(1) - \hat \mu_0(X_i)& &\mbox{and} &D_i^0 \define \hat \mu_1(X_i) - Y_i(0).& 
\end{align*}
In the second stage, estimators for the CATE are derived by regressing the features $X$ on the imputed treatment effects. \cite{kunzel2017meta} provides details. In the X-learner, the estimators of the first stage are held fixed and are not updated in the second stage. This is necessary since, unlike neural networks, many machine learning algorithms, such as RF and BART, cannot be updated in a meaningful way once they have been trained. For neural networks and similar gradient optimization-based algorithms, it is possible to jointly update the estimators in the first and the second stage. 

This is exactly the motivation of the Y-learner. Instead of first deriving an estimator for the control response functions and then an estimator for the CATE function, these functions are optimized jointly. The pseudo-code in Algorithm \ref{algo:Ylearner} shows how these two stages are updated simultaneously. 
In Figure \ref{fig:compareXY}, we compare Y-NN with X-NN, and we find that Y-NN outperforms X-NN for our data sets.

	\clearpage
    \newcommand{\X}{\mathcal{X}}
\newcommand{\I}{\mathcal{I}}
\newcommand{\J}{\mathcal{J}}
\renewcommand{\P}{\mathcal{P}}
\newcommand{\Yobst}{Y^{obs,t}}
\newcommand{\Yobsc}{Y^{obs,c}}
\newcommand{\var}{\mbox{Var}}
\newcommand{\sd}{\mbox{Sd}}
\newcommand{\cov}{\mbox{Cov}}
\newcommand{\Dt}{\tilde D}
\newcommand{\taun}{\hat{\tau}^{mn}}
\newcommand{\theoremfont}{\itshape}
\renewcommand{\descriptionlabel}[1]{\hspace{\labelsep}\textbf{\descriptionlabelfont #1}}
\newcommand{\D}{\mathcal{D}}
\newcommand{\trace}{\mbox{trace}}
\newcommand{\IMSE}{\mbox{IMSE}}
\newcommand{\EMSE}{\mbox{EMSE}}
\newcommand{\MSE}{\mbox{MSE}}
\newcommand{\BIAS}{\mbox{BIAS}}
\newcommand{\mae}{\gamma_{\scriptsize\mbox{max}}}
\newcommand{\mie}{\gamma_{\scriptsize\mbox{min}}}
\newcommand{\mis}{s_{\scriptsize\mbox{min}}}
\newcommand{\ED}{\mathbb{E}_{\mathcal{D}}}
\newcommand{\Ex}{\mathbb{E}_{x}}
\newcommand{\Pobs}{\mathcal{P}_{\mbox{\tiny observed}}}
\newcommand{\deltat}{\tilde \delta}
\newcommand{\Sb}{{\bar S}}
\section{Pseudo Code for CATE Estimators}
\label{app:learners}

In this section, we will present pseudo code for the CATE estimators in this paper. We present code for the meta learning algorithms in Section \ref{app:transfer.algos}.
We denote by $Y^0$ and $Y^1$ the observed outcomes for the control and the treated group. For example, $Y^1_i$ is the observed outcome of the $i$th unit in the treated group. $X^0$ and $X^1$ are the features of the control and treated units, and hence, $X^1_i$  corresponds to the feature vector of the $i$th unit in the treated group. $M_k(Y\sim X)$ is  the notation for a regression estimator, which estimates $x \mapsto \E[Y|X=x]$. It can be any regression/machine learning estimator, but in this paper we only choose it to be a neural network or random forest.

\begin{algorithm}[H] 
	\begin{algorithmic}[1]
		\Procedure{T--Learner}{$X, \Yobs, W$}
		
		\State $\hat \mu_0 	= M_0(Y^0 \sim X^0)$
		\State $\hat \mu_1	 = M_1(Y^1 \sim X^1)$
		
		\item[]
		\State $\hat \tau(x) = \hat \mu_1(x) - \hat \mu_0(x)$ 
		\EndProcedure
	\end{algorithmic}
	\caption{T-learner}
	
	\label{algo:Tlearner}
		\algcomment{$M_0$ and $M_1$ are here some, possibly different machine learning/regression algorithms.}	
\end{algorithm}

\begin{algorithm}[H] 
	\begin{algorithmic}[1]
		\Procedure{S--Learner}{$X, \Yobs, W$}
		
		\State $\hat \mu 	= M(\Yobs \sim (X, W))$
						
		\State $\hat \tau(x) = \hat \mu(x,1) - \hat \mu(x,0)$ 
		\EndProcedure
	\end{algorithmic}
	\caption{S-learner}
	
	\label{algo:Slearner}
	\algcomment{$M(\Yobs \sim (X, W))$ is the notation for estimating $(x,w) \mapsto \E[Y|X=x, W=w]$ while treating $W$ as a 0,1--valued feature.}
\end{algorithm}

\begin{algorithm}[H]
	\begin{algorithmic}[1]	
		\Procedure{X--Learner}{$X, \Yobs, W, g$}

		\item[]
		\State $\hat \mu_0	= M_1(Y^0 \sim X^0)$ 
				\Comment{Estimate response function}
		\State $\hat \mu_1	 = M_2(Y^1 \sim X^1)$
		
		\item[]
		\State $\Dt^1_{i} =         Y^1_{i}                         -     \hat \mu_0(X^1_{i})$
				\Comment{Compute imputed treatment effects}
		\State $\Dt^0_{i} =     \hat \mu_1(X^0_{i})     -     Y^0_{i}$
		
		\item[]
		\State $\hat \tau_1      = M_3(\Dt^{1}      \sim     X^1)$
				\Comment{Estimate CATE for treated and control}
		\State $\hat \tau_0     = M_4(\Dt^0     \sim     X^0)$
		
		\item[]
		\State $\hat \tau(x) = g(x) \hat \tau_0(x) + (1- g(x)) \hat \tau_1(x)$ 
				\Comment{Average the estimates}
		
		\EndProcedure
	\end{algorithmic}
	\caption[test]{X--learner}

	\label{aglo:Xlearner}
	
	\algcomment{$g(x) \in [0,1]$ is a weighing function which is chosen to minimize the variance of $\hat \tau(x)$. It is sometimes possible to estimate $\cov(\tau_0(x) , \tau_1(x) )$, and compute the best $g$ based on this estimate. However, we have made good experiences by choosing $g$ to be an estimate of the propensity score, but also choosing it to be constant and equal to the ratio of treated units usually leads to a good estimator of the CATE. }	
\end{algorithm}

\begin{algorithm}[t]
  \begin{algorithmic}[1]
  \If{$W_i == 0$}
  \State Update the network $\pi_{\theta_0}$ to predict $Y_i^{obs}$
  \State Update the network $\pi_{\theta_1}$ to predict $Y_i^{obs} + \pi_\tau(X_i)$
  \State Update the network $\pi_\tau$ to predict $\pi_{\theta_1}(X_i) - Y_i^{obs}$
  \EndIf
  \If{$W_i == 1$}
  \State Update the network $\pi_{\theta_0}$ to predict $Y_i^{obs} - \pi_\tau(X_i)$
  \State Update the network $\pi_{\theta_1}$ to predict $Y_i^{obs}$
  \State Update the network $\pi_\tau$ to predict $Y_i^{obs} - \pi_{\theta_0}(X_i)$
  \EndIf
  \end{algorithmic}
  \caption{Y-Learner Pseudo Code}
  \label{algo:Ylearner}
  \label{ylearner}
      \algcomment{This process describes training the Y-Learner for one step given a data point $(Y_i^{obs}, X_i, W_i)$}
\end{algorithm}

	\section{Explicit Transfer Learning Algorithms for CATE Estimation}
\label{app:transfer.algos}


\subsection{MLRW Transfer for CATE Estimation}


\begin{algorithm}[H]
	\caption{MLRW Transfer for Cate Estimation.}
	\label{alg:sf-reptile-meta-regression}
	\begin{algorithmic}[1]
		\State Let $\mu_0^{(i)}$ and $\mu_1^{(i)}$ be the outcome under treatment and control for experiment $i$. 
		\State Let numexps be the number of experiments. 
		\State Let $\pi_\theta$ be an $N$ layer neural network parameterized by $\theta = \left[\theta_0, \dots, \theta_{N} \right]$.
		\State Let $\epsilon = \left[\epsilon_0, \dots, \epsilon_{N} \right]$ be a vector, where $N$ is the number of layers in $\pi_\theta$. 
		\State Let outeriters be the total number of training iterations. 
		\State Let inneriters be the number of inner loop training iterations.
		\For{oiter $<$ outeriters}
		\For{i $<$ numexps}
		\State Sample $X_0$ and $X_1$: control and treatment units from experiment $i$
		\For{$j = [0, 1]$} \Comment{j iterating over treatment and control}
		\State Let $U_0(\theta) = \theta$
		\For{k < inneriters}
		\State $\mathcal{L} = \| \pi_{U_k(\theta)} (X_j) - \mu_j(X_j) \|$
		\State Compute $\nabla_\theta \mathcal{L}$. 
		\State Use ADAM with $\nabla_\theta \mathcal{L}$ to obtain $U_{k+1}(\theta)$. 
		\State $U_k (\theta) = U_{k+1}(\theta)$
		\EndFor
		\For{ $p < N$ }
		\State $\theta_p = \epsilon_p \cdot U_k(\theta_p) + (1 - \epsilon_p) \cdot \theta_p$.  
		\EndFor
		\EndFor
		\EndFor
		\EndFor
		\State To Evaluate CATE estimate, \textbf{do}
		\State $C = []$
		\For{i $<$ numexps}
		\State Sample $X_0$ and $X_1$: control and treatment units from experiment $i$
		\State Sample $X$: test units from experiment $i$. 
		\For{$j$ = [0, 1]} \Comment{j iterating over treatment and control}
		\For{$k <$ innteriters}
		\State $\mathcal{L} = \| \pi_{U_k(\theta)} (X_j) - \mu_j(X_j) \|$
		\State Compute $\nabla_\theta \mathcal{L}$. 
		\State Use ADAM with $\nabla_\theta \mathcal{L}$ to obtain $U_{k+1}(\theta)$. 
		\State $U_k (\theta) = U_{k+1}(\theta)$
		\EndFor
		\State $\hat{\mu}_j = \pi_{U_k(\theta)} (X)$
		\EndFor
		\State  $\hat{\tau}_i = \hat{\mu}_0 - \hat{\mu_1}$
		\State C.append$(\hat{\tau}_i)$
		\EndFor
		\State {\bf return} $C$
	\end{algorithmic}
\end{algorithm}

\clearpage
\subsection{Joint Training}
\FloatBarrier
\begin{figure}
	\begin{minipage}[c]{0.45\textwidth}
	\includegraphics[width=0.8\linewidth]{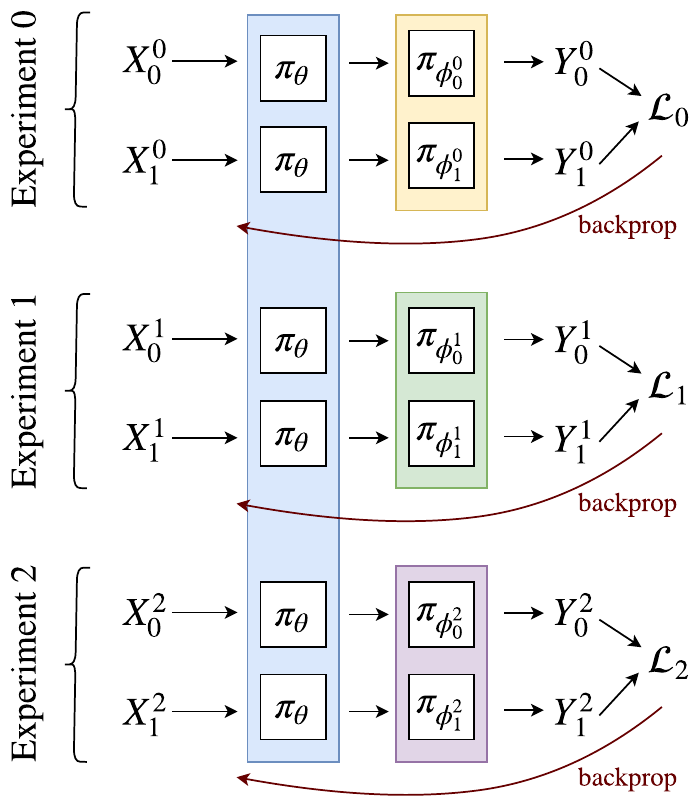}
	\end{minipage}\hfill
	\begin{minipage}[c]{0.4\textwidth}
		\caption{Joint Training - Unlike the Multi-head method which differentiates base layers for treatment and control, the Joint Training method has all observations and experiments (regardless of treatment and control) share the same base network, which extracts general low level features from the data.} \label{fig:joint_training}
	\end{minipage}
\end{figure}

\begin{algorithm}[h]
	\begin{algorithmic}[1]
		\State Let $\mu_0^{(i)}$ and $\mu_1^{(i)}$ be the outcome under control and treatment for experiment $i$. 
		\State Let numexps be the number of experiments. 
		\State Let $\pi_\rho$ be a generic expression for a neural network parameterized by $\rho$. 
		\State Let $\theta$ be base neural network layers shared by all experiments. 
		\State Let $\phi_0^{(i)}$  be neural network layers predicting $\mu_0^{(i)}$ in experiment $i$.
		\State Let $\phi_1^{(i)}$  be neural network layers predicting $\mu_1^{(i)}$ in experiment $i$.
		\State Let $\omega_0^{(i)} = \left[\theta, \phi_0^{(i)} \right]$ be the full prediction network for $\mu_0$ in experiment $i$.
		\State Let $\omega_1^{(i)} = \left[\theta, \phi_1^{(i)} \right]$ be the full prediction network for $\mu_1$ in experiment $i$.
		\State Let $\Omega = \bigcup_{j=0}^{\text{1}} \bigcup_{i=1}^{\text{numexps}} \omega_j^{(i)}$ be all trainable parameters. 
		\State Let numiters be the total number of training iterations
		\For{iter $<$ numiters}
		\State $\mathcal{L}$ = 0
		\For{i $<$ numexps}
		\State Sample $X_0$ and $X_1$: control and treatment units from experiment $i$
		\For{$j$ = [0, 1]} \Comment{j iterating over treatment and control}
		\State $\mathcal{L}_j^{(i)} = \| \pi_{\omega_j^{(i)}} (X_j) - \mu_j(X_j) \|$
		\State $\mathcal{L} = \mathcal{L} + \mathcal{L}_j^{(i)}$
		\EndFor
		\EndFor
		\State Compute $\nabla_\Omega \mathcal{L} = \frac{\partial \mathcal{L}}{\partial \Omega} = \sum_i \sum_j \frac{\partial \mathcal{L}_j^{(i)}}{\partial \omega_j^{(i)}}$
		\State Apply ADAM with gradients given by $\nabla_\Omega \mathcal{L}$. 
		\For{i $<$ numexps}
		\State Sample $X$: test units from experiment $i$
		\EndFor
		\EndFor
		\State $\hat{\mu}_0 = \pi_{\omega_0^{(i)}} (X)$
		\State $\hat{\mu}_1 = \pi_{\omega_1^{(i)}}(X)$
		\State {\bf return} CATE estimate $\hat{\tau} = \hat{\mu}_1 - \hat{\mu}_0$
	\end{algorithmic}
	\caption{Joint Training}
\label{alg:joint-training}
\end{algorithm}

\FloatBarrier

\clearpage
\subsection{T-learner Transfer CATE Estimators}
Here, we present full pseudo code for the algorithms from Section 3 using the T-learner as a base learner. All of these algorithms can be extended to other learners including $S, R, X,$ and $Y$. See the released code for implementations. 

\begin{algorithm}[H]
	\caption{Vanilla T-learner (also referred to as Baseline T-learner)}
	\label{alg:vanilla-t}
	\begin{algorithmic}[1]
		\State Let $\mu_0$ and $\mu_1$ be the outcome under treatment and control. 
		\State Let $X$ be the experimental data. Let $X_t$ be the test data. 
		\State Let $\pi_{\theta_0}$ and $\pi_{\theta_1}$ be a neural networks parameterized by $\theta_0$ and $\theta_1$. 
		\State Let $\theta = \theta_0 \cup \theta_1$. 
		\State Let numiters be the total number of training iterations.
		\State Let batchsize be the number of units sampled. We use 64. 
		\For{i $<$ numiters}
		\State Sample $X_0$ and $X_1$: control and treatment units. Sample batchsize units.  
		\State $\mathcal{L}_0= \| \pi_\theta (X_0) - \mu_0(X_0) \| $
		\State $\mathcal{L}_1 = \| \pi_\theta (X_1) - \mu_1(X_1) \| $
		\State $\mathcal{L} = \mathcal{L}_0 + \mathcal{L}_1$
		\State Compute $\nabla_{\theta} \mathcal{L} = \frac{\partial \mathcal{L}}{\partial \theta}$.
		\State Apply ADAM with gradients given by $\nabla_{\theta} \mathcal{L}$. 
		\EndFor
		\State $\hat{\mu}_0 = \pi_{\theta_0} (X_t)$
		\State $\hat{\mu}_1 = \pi_{\theta_1} (X_t)$
		\State {\bf return} CATE estimate $\hat{\tau} = \hat{\mu}_1 - \hat{\mu}_0$
	\end{algorithmic}
\end{algorithm}

\begin{algorithm}[H]
	\caption{Warm Start T-learner}
	\label{alg:warm-t}
	\begin{algorithmic}[1]
		\State Let $\mu^{i}_0$ and $\mu^{i}_1$ be the outcome under treatment and control for experiment $i$. 
		\State Let $X^i$ be the data for experiment $i$. Let $X^i_t$ be the test data for experiment $i$. 
		\State Let $\pi_{\theta_0}$ and $\pi_{\theta_1}$ be a neural networks parameterized by $\theta_0$ and $\theta_1$. 
		\State Let $\theta = \theta_0 \cup \theta_1$. 
		\State Let numiters be the total number of training iterations.
		\State Let batchsize be the number of units sampled. We use 64. 
		\For{i $<$ numiters}
		\State Sample $X^0_0$ and $X^0_1$: control and treatment units for experiment $0$. Sample batchsize units.  
		\State $\mathcal{L}_0= \| \pi_{\theta_0} (X^0_0) - \mu_0(X^0_0) \| $
		\State $\mathcal{L}_1 = \| \pi_{\theta_1} (X^0_1) - \mu_1(X^0_1) \| $
		\State $\mathcal{L} = \mathcal{L}_0 + \mathcal{L}_1$
		\State Compute $\nabla_{\theta} \mathcal{L} = \frac{\partial \mathcal{L}}{\partial \theta}$.
		\State Apply ADAM with gradients given by $\nabla_{\theta} \mathcal{L}$. 
		\EndFor
		\For{i $<$ numiters}
		\State Sample $X^1_0$ and $X^1_1$: control and treatment units for experiment $1$. Sample batchsize units.  
		\State $\mathcal{L}_0= \| \pi_{\theta_0} (X^1_0) - \mu_0(X^1_0) \| $
		\State $\mathcal{L}_1 = \| \pi_{\theta_1} (X^1_1) - \mu_1(X^1_1) \| $
		\State $\mathcal{L} = \mathcal{L}_0 + \mathcal{L}_1$
		\State Compute $\nabla_{\theta} \mathcal{L} = \frac{\partial \mathcal{L}}{\partial \theta}$.
		\State Apply ADAM with gradients given by $\nabla_{\theta} \mathcal{L}$. 
		\EndFor
		\State $\hat{\mu}_0 = \pi_{\theta_0} (X^1_t)$
		\State $\hat{\mu}_1 = \pi_{\theta_1} (X^1_t)$
		\State {\bf return} CATE estimate $\hat{\tau} = \hat{\mu}_1 - \hat{\mu}_0$
	\end{algorithmic}
\end{algorithm}

\newpage

\begin{algorithm}[H]
	\caption{Frozen Features T-learner}
	\label{alg:frozen-t}
	\begin{algorithmic}[1]
		\State Let $\mu^{i}_0$ and $\mu^{i}_1$ be the outcome under treatment and control for experiment $i$. 
		\State Let $X^i$ be the data for experiment $i$. Let $X^i_t$ be the test data for experiment $i$. 
		\State Let $\pi_\rho$ be a generic expression for a neural network parameterized by $\rho$. 
		\State Let $\theta_0^0, \theta_0^1, \theta_1^0, \theta_1^1$ be neural network parameters. The subscript indicates the outcome that $\theta$ is associated with predicting (0 for control and 1 for treatment) and the superscript indexes the experiment.
		\State Let $\gamma_0$ be the first $k$ layers of $\pi_{\theta^0_0}$. Define $\gamma_1$ analogously. 
		\State Let $\theta^i = \theta^i_0 \cup \theta^i_1$. 
		\State Let numiters be the total number of training iterations.
		\State Let batchsize be the number of units sampled. We use 64. 
		\For{i $<$ numiters}
		\State Sample $X^0_0$ and $X^0_1$: control and treatment units for experiment $0$. Sample batchsize units.  
		\State $\mathcal{L}_0= \| \pi_{\theta^0_0} (X^0_0) - \mu_0(X^0_0) \| $
		\State $\mathcal{L}_1 = \| \pi_{\theta^0_1} (X^0_1) - \mu_1(X^0_1) \| $
		\State $\mathcal{L} = \mathcal{L}_0 + \mathcal{L}_1$
		\State Compute $\nabla_{\theta} \mathcal{L} = \frac{\partial \mathcal{L}}{\partial \theta}$.
		\State Apply ADAM with gradients given by $\nabla_{\theta^0} \mathcal{L}$. 
		\EndFor
		\For{i $<$ numiters}
		\State Sample $X^1_0$ and $X^1_1$: control and treatment units for experiment $1$. Sample batchsize units. 
		\State Compute $Z^1_0 = \pi_\gamma(X^1_0)$ and $Z^1_1 = \pi_\gamma(X^1_1)$
		\State $\mathcal{L}_0= \| \pi_{\theta_0^1} (Z^1_0) - \mu_0(Z^1_0) \| $
		\State $\mathcal{L}_1 = \| \pi_{\theta_1^1} (Z^1_1) - \mu_1(Z^1_1) \| $
		\State $\mathcal{L} = \mathcal{L}_0 + \mathcal{L}_1$
		\State Compute $\nabla_{\theta^1} \mathcal{L} = \frac{\partial \mathcal{L}}{\partial \theta^1}$. Do not compute gradients with respect to $\theta^0$ parameters. 
		\State Apply ADAM with gradients given by $\nabla_{\theta^1} \mathcal{L}$. 
		\EndFor
		\State Compute $Z^1_t = \pi_\gamma(X^1_t)$. 
		\State $\hat{\mu_0} = \pi_{\theta^1_0} (Z^1_t)$
		\State $\hat{\mu_1} = \pi_{\theta^1_1} (Z^1_t)$
		\State {\bf return} CATE estimate $\hat{\tau} = \hat{\mu_1} - \hat{\mu_0}$
	\end{algorithmic}
\end{algorithm}

\newpage
\newpage 
\begin{algorithm}[H]
	\caption{Multi-Head T-learner}
	\label{alg:multi-t}
	\begin{algorithmic}[1]
		\State Let $\mu^{i}_0$ and $\mu^{i}_1$ be the outcome under treatment and control for experiment $i$. 
		\State Let $X^i$ be the data for experiment $i$. Let $X^i_t$ be the test data for experiment $i$. 
		\State Let $\pi_\rho$ be a generic expression for a neural network parameterized by $\rho$. 
		\State Let $\theta_0$ be base neural network layers shared by all experiments for predicting outcomes under control.
		\State Let $\theta_1$ be base neural network layers shared by all experiments for predicting outcomes under treatment.
		\State Let $\phi_0^{(i)}$  be neural network layers receiving $\pi_{\theta_0}(x^i_0)$ as input and predicting $\mu_0^{(i)}(x^i_0)$ in experiment $i$.
		\State Let $\phi_1^{(i)}$  be neural network layers receiving $\pi_{\theta_1}(x^i_1)$ as input and predicting $\mu_1^{(i)}(x^i_1)$ in experiment $i$.
		\State Let $\omega_0^{(i)} = \left[\theta, \phi_0^{(i)} \right]$ be all trainable parameters used to predict $\mu_0^i$.
		\State Let $\omega_1^{(i)} = \left[\theta, \phi_1^{(i)} \right]$ be all trainable parameters used to predict $\mu_1^i$.
		\State Let $\Omega^i = \omega_0^{(i)} \cup \omega_1^{(i)}$. 
		\State Let numiters be the total number of training iterations.
		\State Let batchsize be the number of units sampled. We use 64. 
		\State Let numexps be the number of experiments. 
		\For{i $<$ numiters}
		\For{j $<$ numexps}
		\State Sample $X^j_0$ and $X^j_1$: control and treatment units for experiment $j$. Sample batchsize units.  
		\State Compute $Z_0^j = \pi_{\theta_0}(X^j_0)$ and $Z_1^j = \pi_{\theta_1}(X^j_1)$
		\State Compute $\hat{\mu}_0^j = \pi_{\phi_0^j} (z_0^j)$ and $\hat{\mu}_1^j = \pi_{\phi_1^j} (z_1^j)$
		\State $\mathcal{L}_0= \| \hat{\mu}_0^j - \mu_0^j(X^j_0) \| $
		\State $\mathcal{L}_1 = \| \hat{\mu}_1^j - \mu_1^j(X^j_1) \| $
		\State $\mathcal{L} = \mathcal{L}_0 + \mathcal{L}_1$
		\State Compute $\nabla_{\Omega^i} \mathcal{L} = \frac{\partial \mathcal{L}}{\partial \Omega^i}$.
		\State Apply ADAM with gradients given by $\nabla_{\theta} \mathcal{L}$. 
		\EndFor
		\EndFor
		\State Let $C = []$
		\For{j $<$ numexps}
		\State Compute $Z_0^j = \pi_{\theta_0}(X^j_t)$ and $Z_1^j = \pi_{\theta_1}(X^j_t)$
		\State Compute $\hat{\mu}_0^j = \pi_{\phi_0^j} (z_0^j)$ and $\hat{\mu}_1^j = \pi_{\phi_1^j} (z_1^j)$
		\State Estimate CATE $\hat{\tau} = \hat{\mu}_1^j - \hat{\mu}_0^j$. 
		\State $C.$append$(\hat{\tau})$
		\EndFor
		\State {\bf return} $C$ 
	\end{algorithmic}
\end{algorithm}

\newpage

\begin{algorithm}[H]
	\caption{SF Reptile T-learner}
	\label{alg:sf-reptile-t}
	\begin{algorithmic}[1]
		\State Let $\mu_0^i$ and $\mu_1^i$ be the outcome under treatment and control for experiment $i$. 
		\State Let $X^i$ be the data for experiment $i$. Let $X^i_t$ be the test data for experiment $i$. 
		\State Let $\pi_{\theta_0}$ and $\pi_{\theta_1}$ be a neural networks parameterized by $\theta_0$ and $\theta_1$. 
		\State Let $\theta = \theta_0 \cup \theta_1$. 
		\State Let $\epsilon = \left[\epsilon_0, \dots, \epsilon_{N} \right]$ be a vector, where $N$ is the number of layers in $\pi_{\theta_i}$. 
		\State Let numouteriters be the total number of outer training iterations.
		\State Let numinneriters be the total number of inner training iterations. 
		\State Let numexps be the number of experiments. 
		\State Let batchsize be the number of units sampled. We use 64. 
		\For{iouter $<$ numouteriters}
		\For{i $<$ numexps}
		\State $U_0(\theta_0) = \theta_0$ 
		\State $U_0(\theta_1) = \theta_1$. 
		\For{k$<$ numinneriters}
		\State Sample $X_0^i$ and $X_1^i$: control and treatment units. Sample batchsize units.  
		\State $\mathcal{L}_0= \| \pi_{U_k(\theta_0)} (X_0^i) - \mu_0(X_0^i) \| $
		\State $\mathcal{L}_1 = \| \pi_{U_k(\theta_1)} (X_1^i) - \mu_1(X_1^i) \| $
		\State $\mathcal{L} = \mathcal{L}_0 + \mathcal{L}_1$
		\State Compute $\nabla_{\theta} \mathcal{L} = \frac{\partial \mathcal{L}}{\partial \theta}$.
		\State Use ADAM with gradients given by $\nabla_{\theta} \mathcal{L}$ to obtain $U_{k+1}(\theta_0)$ and $U_{k+1}(\theta_1)$. 
		\State Set $U_{k}(\theta_0)=U_{k+1}(\theta_0)$ and $U_{k}(\theta_1) = U_{k+1}(\theta_1)$
		\EndFor
		\For{ $p < N$ }
		\State $\theta_p = \epsilon_p \cdot U_k(\theta_p) + (1 - \epsilon_p) \cdot \theta_p$.  
		\EndFor
		\EndFor
		\EndFor
		\State To Evaluate CATE estimate, \textbf{do}
		\State $C = []$. 
		\For{i $<$ numexps}
		\State $U_0(\theta_0) = \theta_0$ 
		\State $U_0(\theta_1) = \theta_1$. 
		\For{k$<$ numinneriters}
		\State Sample $X_0^i$ and $X_1^i$: control and treatment units. Sample batchsize units.  
		\State $\mathcal{L}_0= \| \pi_{U_k(\theta_0)} (X_0^i) - \mu_0(X_0^i) \| $
		\State $\mathcal{L}_1 = \| \pi_{U_k(\theta_1)} (X_1^i) - \mu_1(X_1^i) \| $
		\State $\mathcal{L} = \mathcal{L}_0 + \mathcal{L}_1$
		\State Compute $\nabla_{\theta} \mathcal{L} = \frac{\partial \mathcal{L}}{\partial \theta}$.
		\State Use ADAM with gradients given by $\nabla_{\theta} \mathcal{L}$ to obtain $U_{k+1}(\theta_0)$ and $U_{k+1}(\theta_1)$. 
		\State Set $U_{k}(\theta_0)=U_{k+1}(\theta_0)$ and $U_{k}(\theta_1) = U_{k+1}(\theta_1)$
		\EndFor
		\State $\hat{\mu}_0^i = \pi_{U_k(\theta_0)}(X_0^i)$
		\State $\hat{\mu}_1^i = \pi_{U_k(\theta_1)}(X_1^i)$
		\State $\hat{\tau}^i =  \hat{\mu}_1^i - \hat{\mu}_0^i$
		\State $C.$append$(\hat{\tau}^i)$. 
		\EndFor
		\State {\bf return} $C$. 
	\end{algorithmic}
\end{algorithm}

    \clearpage
\begin{sidewaystable}[ht]
\begin{flushleft}
\begin{tabular}{rrrrrrrrrrrrrrrr}
 {} & {\textbf{Method}} & {\textbf{LM-1}} & {\textbf{LM-2}} & {\textbf{LM-3}} & {\textbf{LM-4}} & {\textbf{LM-5}} & {\textbf{LM-6}} & {\textbf{LM-7}} & {\textbf{LM-8}} & {\textbf{LM-9}} & {\textbf{LM-10}} & {\textbf{LM-11}} & {\textbf{LM-12}} & {\textbf{LM-13}} & {\textbf{LM-14}} \\ 
  \hline
{\textbf{t-lm}} &  & 15.95 & 7.82 & 20.14 & 6.62 & 46.15 & 17.24 & 9.88 & 10.65 & 44.77 & 7.63 & 8.43 & 9 & 11.21 & 20.9 \\ 
   \hline
{\textbf{s-rf}} &  & 19.13 & 13.18 & 7.62 & 13.27 & 15.66 & 15.58 & 11.44 & 16.3 & 11.34 & 16.53 & 12.57 & 14.11 & 13.49 & 18.56 \\ 
   \hline
{\textbf{t-rf}} &  & 20.04 & 13.56 & 7.79 & 13.62 & 16 & 15.99 & 11.76 & 16.65 & 11.54 & 17.11 & 13.66 & 14.38 & 13.67 & 18.96 \\ 
   \hline
{\textbf{R-NN}} &  & 18.95 & 5.19 & 9.33 & 28.98 & 3.04 & 10.03 & 4.88 & 16.34 & 7.91 & 14.56 & 19.23 & 3.5 & 10.4 & 15.53 \\ 
   \hline
{\textbf{S-NN}} & frozen & 6.74 & 6.17 & 2.75 & 5.76 & 5.68 & 6.27 & 4.23 & 7.17 & 4.15 & 6.77 & 4.88 & 5.9 & 6.63 & 9.87 \\ 
   & multi head & 3.3 & 3.05 & 2.34 & 3.83 & 3.6 & 4.89 & 4.56 & 8.47 & 5.87 & 5.22 & 5.43 & 6.77 & 5.15 & 5.58 \\ 
   & SF & 4.85 & 38.65 & 7.54 & 55.92 & 11.36 & 3.98 & 50.76 & 7.74 & 5.72 & 7.73 & 8.64 & 31.06 & 30.81 & 18.08 \\ 
   & baseline & 6.84 & 5.29 & 3.77 & 6.86 & 5.94 & 7.19 & 4.6 & 8.08 & 5.61 & 7.12 & 4.34 & 6.05 & 8.59 & 10.75 \\ 
   & warm & 7.29 & 6.44 & 4.08 & 7.25 & 6.74 & 7.17 & 6.03 & 8.8 & 5.63 & 7.6 & 5.82 & 6.53 & 8.55 & 12.02 \\ 
   \hline
{\textbf{T-NN}} & frozen & 6.53 & 6.73 & 4.49 & 6.35 & 7.23 & 6.61 & 5.99 & 7.59 & 5.79 & 6.79 & 5.99 & 6.58 & 7.38 & 9.39 \\ 
   & multi head & 2.73 & 2.34 & 1.34 & 2.11 & 2.49 & 2.3 & 1.97 & 2.73 & 1.94 & 2.36 & 1.95 & 2.32 & 2.65 & 3.64 \\ 
   & SF & 20.72 & 27.71 & 8.54 & 20.18 & 23.06 & 15.35 & 14.27 & 13.05 & 9.37 & 27.94 & 40.03 & 16.3 & 20.81 & 6.78 \\ 
   & baseline & 22.76 & 5.34 & 5.98 & 5.84 & 5.16 & 10.37 & 10.9 & 7.26 & 10.25 & 10.18 & 6.26 & 5.69 & 6.71 & 10.31 \\ 
   & warm & 23.46 & 7.2 & 5.75 & 6.41 & 5.21 & 11.56 & 12.21 & 8.77 & 8.93 & 6.81 & 8.93 & 6.15 & 7.25 & 18.98 \\ 
   \hline
{\textbf{X-NN}} & frozen & 6.63 & 14.04 & 10.73 & 19.57 & 17.94 & 14 & 13.67 & 18.83 & 14.36 & 10.81 & 12.25 & 16.61 & 39.96 & 32.81 \\ 
   & multi head & 1.19 & 11.38 & 84.13 & 174.18 & 1.87 & 19.55 & 62.12 & 22.7 & 9.67 & {\textbf{0.85*}} & 3.34 & 5.03 & {\textbf{0.94*}} & 111.42 \\ 
   & SF & 18.83 & 10.72 & 10.3 & 10.61 & 10.11 & 12.5 & 21.37 & 11.12 & 8.27 & 16.33 & 9.81 & 13.8 & 9.85 & 10.15 \\ 
   & baseline & 19.52 & 8.25 & 4.68 & 5.06 & 6.6 & 11.5 & 10.78 & 12 & 10.26 & 6.25 & 13.11 & 7.27 & 9.2 & 18.45 \\ 
   & warm & 20.06 & 8.54 & 5.57 & 6.37 & 10.77 & 9.34 & 11.36 & 13.16 & 8.03 & 8.66 & 10.96 & 7.52 & 11.57 & 16.7 \\ 
   \hline
{\textbf{Y-NN}} & frozen & 1.54 & {\textbf{1*}} & 2.1 & 3.3 & {\textbf{1.11*}} & 46.07 & 27.63 & 5.47 & 7.48 & 7.21 & 1.02 & {\textbf{1.15*}} & 0.97 & 43.82 \\ 
   & multi head & 0.92 & 2.21 & 1.26 & 15.86 & 1.19 & 20.47 & 1.76 & 2.75 & 3.96 & 9.4 & 19.11 & 1.26 & 35.43 & 9.29 \\ 
   & SF & 5.68 & 12 & 16.57 & 38 & 5.54 & 3.49 & 8.2 & 4.8 & 2.61 & 7.21 & 8.48 & 12.98 & 9.58 & 5.08 \\ 
   & baseline & {\textbf{0.9*}} & 1.31 & 5.24 & 31.43 & 6.8 & {\textbf{1.23*}} & 7.5 & {\textbf{1.07*}} & 1.32 & 1.35 & 8.19 & 1.2 & 4.11 & 7.72 \\ 
   & warm & 48.91 & 1.26 & 5.79 & 3.61 & 29.43 & 2.71 & 3.94 & 12.87 & 21.76 & 13.25 & 15.78 & 17.45 & 1.12 & 29.01 \\ 
   \hline
{\textbf{joint}} & joint & 13.75 & 5.81 & 3.46 & 4.12 & 3.46 & 12.22 & 9.58 & 14.96 &  & 4.75 & 8.55 & 13.13 & 9.91 & 11.68 \\ 
   \hline
{\textbf{MLRW}} &  & 1 & 1.41 & {\textbf{1.05*}} & {\textbf{1.94*}} & 1.97 & 1.26 & {\textbf{1.03*}} & 2.05 & {\textbf{0.9*}} & 1.85 & {\textbf{1*}} & 1.15 & 1.57 & {\textbf{2.75*}} \\ 
  \end{tabular}
\caption{MSE in percent for different CATE estimators.} 
\label{table:LM}
\end{flushleft}
\end{sidewaystable}

\begin{sidewaystable}[ht]
\begin{flushleft}
\begin{tabular}{rrrrrrrrrrrrrrrr}
 {} & {\textbf{Method}} & {\textbf{RF-1}} & {\textbf{RF-2}} & {\textbf{RF-3}} & {\textbf{RF-4}} & {\textbf{RF-5}} & {\textbf{RF-6}} & {\textbf{RF-7}} & {\textbf{RF-8}} & {\textbf{RF-9}} & {\textbf{RF-10}} & {\textbf{RF-11}} & {\textbf{RF-12}} & {\textbf{RF-13}} & {\textbf{RF-14}} \\ 
  \hline
{\textbf{t-lm}} &  & 17.95 & 26.73 & 2.48 &  & 4.84 & 1.76 & 6.2 & 7.55 & 24.58 & 3.5 & 2.35 & 2.1 & 3.39 & 7.41 \\ 
   \hline
{\textbf{s-rf}} &  & 7.18 & 7 & 4.18 & 6.86 & 8.43 & 9.15 & 6.03 & 9.28 & 5.95 & 8.49 & 6.7 & 8.1 & 6.46 & 8.64 \\ 
   \hline
{\textbf{t-rf}} &  & 10.95 & 7.76 & 4.69 & 7.79 & 9.03 & 10.07 & 6.67 & 10.08 & 6.3 & 10.35 & 8.19 & 9.28 & 6.73 & 10.04 \\ 
   \hline
{\textbf{R-NN}} &  & 11.41 & 1.18 & 5.7 & 1.26 & 0.7 & 3.79 & 9.11 & 8.01 & 7.17 & 5.14 & 4.15 & 0.52 & 1.26 & 6.88 \\ 
   \hline
{\textbf{S-NN}} & frozen & 1.56 & 0.66 & 0.7 & 0.77 & 0.87 & 0.83 & 0.91 & 0.89 & 0.59 & 0.87 & 0.6 & 0.92 & 0.91 & 1.23 \\ 
   & multi head & 0.69 & 1.23 & 0.4 & 1.59 & 1.58 & 0.65 & 0.81 & 0.57 & {\textbf{0.28*}} & 1.98 & 1.08 & 1.24 & 2.27 & 3.66 \\ 
   & SF & 1.36 & 9.64 & 4.26 & 4.53 & 6 & 6.74 & 11.68 & 2.8 & 12.01 & 41.6 & 45.72 & 4.59 & 1.98 & 6.75 \\ 
   & baseline & 0.91 & 1.28 & 0.84 & 0.93 & 1.78 & 2.05 & 1.16 & 2.04 & 1.61 & 1.06 & 1.29 & 1.49 & 1.99 & 2.69 \\ 
   & warm & 1.08 & 1.3 & 0.94 & 1.16 & 1.85 & 2.22 & 1.4 & 2.15 & 1.65 & 1.12 & 1.37 & 1.42 & 1.83 & 2.52 \\ 
   \hline
{\textbf{T-NN}} & frozen & 1.55 & 1.02 & 0.62 & 0.95 & 1.11 & 0.99 & 0.89 & 1.18 & 0.86 & 1.03 & 0.89 & 1.01 & 1.14 & 1.49 \\ 
   & multi head & 0.66 & 0.58 & 0.35 & 0.53 & 0.63 & 0.53 & 0.47 & 0.63 & 0.49 & 0.57 & 0.51 & 0.55 & 0.64 & 0.91 \\ 
   & SF & 11.66 & 27.5 & 6.89 & 22.59 & 20.99 & 12.04 & 19.77 & 7.85 & 5.39 & 17.31 & 17.15 & 9.1 & 30.42 & 3.55 \\ 
   & baseline & 7.83 & 4.25 & 1.44 & 1.47 & 1.35 & 1.33 & 2.05 & 8.55 & 1.79 & 3.21 & 2.29 & 3.62 & 2.03 & 8.7 \\ 
   & warm & 12.74 & 3.32 & 1.33 & 1.58 & 1.06 & 1.6 & 2.19 & 9.28 & 1.52 & 2.77 & 3.73 & 4.99 & 1.64 & 8.74 \\ 
   \hline
{\textbf{X-NN}} & frozen & 2.53 & 22.89 & 55.57 & 44.11 & 4.18 & 5.72 & 33.18 & 2.62 & 7.59 & 4.96 & 4.41 & 2.01 & 3.01 & 1.23 \\ 
   & multi head & 3.45 & 2.53 & 47.09 & 39.7 & 2 & 27.62 & 10.39 & 0.78 & 11.8 & 10.85 & 0.93 & 9.72 & 11.74 & 30.62 \\ 
   & SF & 4.34 & 1.85 & 5.47 & 6.82 & 4.21 & 11.89 & 7.91 & 5.02 & 4.67 & 6.82 & 16.39 & 6.99 & 5.22 & 2.11 \\ 
   & baseline & 4.09 & 2.05 & 1.67 & 0.73 & 0.58 & 1.16 & 4.97 & 9.23 & 4.05 & 1.49 & 5.17 & 2.15 & 4.07 & 8.05 \\ 
   & warm & 2.51 & 1.73 & 1.72 & 1.48 & 0.97 & 1.04 & 7.81 & 4.35 & 4.59 & 1.34 & 3.86 & 1.37 & 2.08 & 4.83 \\ 
   \hline
{\textbf{Y-NN}} & frozen & 0.42 & {\textbf{0.32*}} & 10.65 & 0.72 & 36 & 1.62 & 0.82 & 0.87 & 15.19 & 1.61 & 0.77 & 46.3 & 28.37 & 1.98 \\ 
   & multi head & 29.41 & 9.71 & 2.74 & 12.27 & 48.87 & 16.59 & 50.21 & 73.74 & 1.65 & 1.76 & 3.52 & 24.26 & 0.76 & 216.69 \\ 
   & SF & 5.45 & 2.51 & 1.53 & 4.76 & 9.61 & 20.81 & 3.68 & 4.84 & 10.8 & 2.87 & 2.78 & 1.36 & 9.07 & 4.08 \\ 
   & baseline & 0.71 & 1.06 & 1.54 & 0.57 & 2.44 & 0.47 & 0.73 & 1.38 & 1.25 & 1.77 & 0.71 & 0.29 & 0.63 & 1.98 \\ 
   & warm & 0.9 & 27.01 & 0.52 & 22.7 & 1.37 & 24.94 & {\textbf{0.3*}} & 2.08 & 12.36 & 3.39 & 9.58 & 2.74 & 2.47 & 11.48 \\ 
   \hline
{\textbf{joint}} & joint &  & 33.47 & 1.3 & 0.61 & 5.15 & 0.47 & 2.68 &  & 2.67 & {\textbf{0.37*}} & 7.76 & 28.14 & 7.48 & 10.04 \\ 
   \hline
{\textbf{MLRW}} &  & {\textbf{0.37*}} & 1.12 & {\textbf{0.18*}} & {\textbf{0.3*}} & {\textbf{0.46*}} & {\textbf{0.21*}} & 0.42 & {\textbf{0.45*}} & 0.37 & 1.18 & {\textbf{0.35*}} & {\textbf{0.23*}} & 0.26 & {\textbf{0.16*}} \\ 
  \end{tabular}
\caption{MSE in percent for different CATE estimators.} 
\label{table:RF}
\end{flushleft}
\end{sidewaystable}

\begin{sidewaystable}[ht]
\begin{flushleft}
\begin{tabular}{rrrrrrrrrrrrrrrr}
 {} & {\textbf{Method}} & {\textbf{RFt-1}} & {\textbf{RFt-2}} & {\textbf{RFt-3}} & {\textbf{RFt-4}} & {\textbf{RFt-5}} & {\textbf{RFt-6}} & {\textbf{RFt-7}} & {\textbf{RFt-8}} & {\textbf{RFt-9}} & {\textbf{RFt-10}} & {\textbf{RFt-11}} & {\textbf{RFt-12}} & {\textbf{RFt-13}} & {\textbf{RFt-14}} \\ 
  \hline
{\textbf{t-lm}} &  & 22.65 & 2.74 & 2.3 & 5.07 & 4.78 & 8.3 & 9.77 & 38.73 & 23.94 & 8.94 & 83.37 & 4.26 & 4.31 & 31.76 \\ 
   \hline
{\textbf{s-rf}} &  & 3.54 & 7.04 & 4.39 & 6.4 & 8.09 & 6.41 & 11.44 & 6.33 & 6.05 & 6.52 & 5.09 & 9.29 & 11.64 & 4.28 \\ 
   \hline
{\textbf{t-rf}} &  & 13.3 & 13.55 & 8.41 & 11.9 & 14.11 & 15.48 & 14.42 & 12.54 & 10.19 & 18.99 & 12.48 & 14.64 & 13.61 & 12.02 \\ 
   \hline
{\textbf{R-NN}} &  & 62.08 & 17.68 & 5.22 & 19.91 & 14.43 & 20.53 & 18.51 & 49.63 & 8.99 & 86.89 & 54.17 & 23.86 & 3.02 & 18.98 \\ 
   \hline
{\textbf{S-NN}} & frozen & 2.49 & 51.39 & 37.47 & 49.5 & 43.17 & 29.19 & 21.81 & 27.25 & 61.34 & 27.02 & 17.25 & 64.22 & 24.94 & 11.24 \\ 
   & multi head & 1.4 & 0.57 & 0.73 & 1.48 & 1.54 & 1.27 & 1.1 & 1.63 & 0.88 & 1.07 & 1.53 & 0.79 & 0.74 & 1.59 \\ 
   & SF & 0.84 & 68.31 & 4.12 & 10.77 & 17.28 & 20.06 & 10.22 & 1.08 & 8.95 & 9.11 & 5.97 & 19.67 & 11.64 & 8.68 \\ 
   & baseline & 14.66 & 2.08 & 0.95 & 1.38 & 2.24 & 7.3 & 3.17 & 1.08 & 0.88 & 5.29 & 1.44 & 2.37 & 2.85 & 1.78 \\ 
   & warm & 17.64 & 2.76 & 1.38 & 2.31 & 1.99 & 8.64 & 4.32 & 1.37 & 0.85 & 5.5 & 3.76 & 5.44 & 3.13 & 3.49 \\ 
   \hline
{\textbf{T-NN}} & frozen & 2.57 & 1.69 & 1.12 & 1.55 & 1.81 & 1.6 & 1.52 & 1.86 & 1.43 & 1.63 & 1.47 & 1.68 & 1.85 & 2.27 \\ 
   & multi head & 0.69 & 0.55 & 0.34 & 0.49 & 0.63 & 0.64 & 0.51 & 0.64 & 0.5 & 0.53 & {\textbf{0.46*}} & {\textbf{0.55*}} & 0.66 & 0.8 \\ 
   & SF & 18.04 & 9.54 & 7.33 & 6.61 & 22.11 & 9.21 & 20.2 & 10.65 & 15.06 & 13.87 & 27.64 & 17.96 & 14.08 & 4.49 \\ 
   & baseline & 43.37 & 14.55 & 9.48 & 19.35 & 15.68 & 20.91 & 16 & 41.72 & 6.99 & 30.02 & 45.59 & 16.03 & 4.79 & 18.63 \\ 
   & warm & 69.41 & 19.81 & 12.99 & 24.12 & 20.28 & 37.07 & 19.64 & 39.37 & 9.36 & 40.65 & 37.59 & 18.62 & 6.87 & 32.52 \\ 
   \hline
{\textbf{X-NN}} & frozen & 12.41 & 221.49 & 215.01 & 131.46 & 94.89 & 328.53 & 199.48 & 41.14 & 398.33 & 83.75 & 132 & 434.25 & 125.88 & 168.92 \\ 
   & multi head & 12.24 & 0.79 & 3.22 & 12.46 & 2.87 & 12.11 & 21.38 & 1.86 & 1.48 & 18.46 & 1.82 & 99.38 & 14.67 & 28.43 \\ 
   & SF & 3.98 & 11.03 & 4.33 & 5.31 & 5.42 & 4.67 & 14.03 & 5.99 & 7 & 7 & 6.42 & 7.21 & 4.89 & 2.34 \\ 
   & baseline & 60.44 & 15.03 & 4.93 & 7.08 & 9.51 & 21.22 & 14.7 & 16.83 & 8 & 80.71 & 23.89 & 25.5 & 4.67 & 17.1 \\ 
   & warm & 30.46 & 13.02 & 5.11 & 9.95 & 9.21 & 13.12 & 9.95 & 12.8 & 4.48 & 50.45 & 19.27 & 12.92 & 8.1 & 15.73 \\ 
   \hline
{\textbf{Y-NN}} & frozen & 1.72 & {\textbf{0.33*}} & 0.87 & 2.47 & 1.8 & 5.44 & 1.35 & 19.37 & 37.98 & 0.81 & 0.7 & 33.11 & 1.34 & 13.64 \\ 
   & multi head & 5.95 & 11.12 & 0.41 & 2.94 & 2.49 & 24.73 & 3.44 & 58.63 & 12.99 & 0.41 & 9.55 & 0.6 & 3.93 & 0.49 \\ 
   & SF & 3.98 & 14.56 & 12.85 & 4.3 & 3.66 & 9.78 & 11.13 & 3.03 & 5.78 & 3.37 & 11.87 & 11.33 & 4.51 & 2.64 \\ 
   & baseline & 0.88 & 10.92 & 0.28 & 1.32 & 1.89 & 11.35 & 8.42 & 1.6 & 5.66 & 0.54 & 0.54 & 2.01 & 0.61 & 2.13 \\ 
   & warm & 8.03 & 0.48 & 0.36 & 0.71 & 2.53 & 1.21 & {\textbf{0.31*}} & 1.77 & {\textbf{0.41*}} & 0.74 & 23.48 & 1.4 & 1.15 & 17.72 \\ 
   \hline
{\textbf{joint}} & joint & 102.83 & 38.47 & 133.88 & 15.31 & 37.32 & 410.04 & 38.28 & 14.96 & 5.02 & 11.97 & 33.71 & 26.33 & 11.99 & 147.45 \\ 
   \hline
{\textbf{MLRW}} &  & {\textbf{0.31*}} & 1.25 & {\textbf{0.21*}} & {\textbf{0.27*}} & {\textbf{0.35*}} & {\textbf{0.29*}} & 0.33 & {\textbf{0.61*}} & 0.61 & {\textbf{0.4*}} & 0.57 & 1.32 & 0.8 & {\textbf{0.24*}} \\ 
  \end{tabular}
\caption{MSE in percent for different CATE estimators.} 
\label{table:RF1}
\end{flushleft}
\end{sidewaystable}

\end{document}